\documentclass[10pt]{article} 
\usepackage[preprint]{tmlr}
\definecolor{iclrgreen}{rgb}{0.0, 0.45, 0.0} 
\usepackage[colorlinks=true,
    citecolor=iclrgreen,
    linkcolor=iclrgreen,
    urlcolor=iclrgreen]{hyperref}
\usepackage{xcolor}
\setcitestyle{authoryear,open={},close={}}

\usepackage{amsmath,amsfonts,bm}

\def\eqref#1{equation~\ref{#1}}

\def\1{\bm{1}}

\DeclareMathAlphabet{\mathsfit}{\encodingdefault}{\sfdefault}{m}{sl}
\SetMathAlphabet{\mathsfit}{bold}{\encodingdefault}{\sfdefault}{bx}{n}

\usepackage{hyperref}
\usepackage{url}
\usepackage{amsthm}
\usepackage{hyphenat}

\usepackage[ruled,vlined,linesnumbered]{algorithm2e}

\usepackage{booktabs}
\newtheorem{proposition}{Proposition}
\newtheorem{lemma}{Lemma}

\newtheorem{corollary}{Corollary}
\usepackage{graphicx}
\usepackage{todonotes}
\usepackage{algpseudocode}
\usepackage{xcolor}
\title{Breaking the Bottlenecks: Scalable Diffusion Models for 3D Molecular Generation}

\author{
\name Adrita Das \email adrita.riman@gmail.com \\
\addr Independent Researcher
\AND
\name Peiran Jiang \email peiranj@andrew.cmu.edu \\
\addr Carnegie Mellon University
\AND
\name Dantong Zhu \email dantongzhu1103@gmail.com \\
\addr Columbia University
\AND
\name Barnabas Poczos \email bapoczos@cs.cmu.edu \\
\addr Carnegie Mellon University
\AND
\name Jose Lugo-Martinez \email jlugomar@andrew.cmu.edu \\
\addr Carnegie Mellon University
}

\def\month{MM}  
\def\year{YYYY} 
\def\openreview{\url{https://openreview.net/forum?id=XXXX}} 

\begin{document}

\maketitle

\begin{abstract}
  Diffusion models have emerged as a powerful class of generative models for molecular design, capable of capturing complex structural distributions and achieving high fidelity in 3D molecule generation. However, their widespread use remains constrained by long sampling trajectories, stochastic variance in the reverse process, and limited structural awareness in denoising dynamics. The Directly Denoising Diffusion Model (DDDM) mitigates these inefficiencies by replacing stochastic reverse MCMC updates with deterministic denoising step, substantially reducing inference time. Yet, the theoretical underpinnings of such deterministic updates have remained opaque. In this work, we provide a principled reinterpretation of DDDM through the lens of the Reverse Transition Kernel (RTK) framework by (\cite{huang2024reversetransitionkernelflexible}), unifying deterministic and stochastic diffusion under a shared probabilistic formalism. By expressing the DDDM reverse process as an approximate kernel operator, we show that the direct denoising process implicitly optimizes a structured transport map between noisy and clean samples. This perspective elucidates why deterministic denoising achieves efficient inference. Beyond theoretical clarity, this reframing resolves several long-standing bottlenecks in molecular diffusion. The RTK view ensures numerical stability by enforcing well-conditioned reverse kernels, improves sample consistency by eliminating stochastic variance, and enables scalable and symmetry-preserving denoisers that respect SE(3) equivariance. Empirically, we demonstrate that RTK-guided deterministic denoising achieves faster convergence and higher structural fidelity than stochastic diffusion models, while preserving chemical validity across GEOM-DRUGS dataset. Code, models, and datasets are publicly available in our
\href{https://github.com/adrita78/breaking-bottlenecks-3d-diffusion/tree/main}{project repository}. 

\textcolor{blue}{\textit{Note: Minor updates are in progress. Please consult the latest version for the updated results.}}

\end{abstract}

\section{Introduction}
Generation of novel, valid molecules is a computationally
intensive task, as navigating the vast chemical space often
requires significant resources to process and model complex
molecular interactions. Recent advances in deep learning
have accelerated this process, but the computational demands of these models remain a challenge, particularly in resource-constrained environments. In the rapidly evolving field of molecular therapeutics and materials discovery, generative models hold immense promise for streamlining key stages of the design process. Diffusion models (\cite{ho2020denoisingdiffusionprobabilisticmodels}; \cite{song2021scorebasedgenerativemodelingstochastic}; \cite{nichol2021improveddenoisingdiffusionprobabilistic}), in particular, have emerged as a leading tool due to their ability to generate diverse high quality samples from learned distributions. These models excel in unconditional generation (\cite{kingma2023variationaldiffusionmodels}) and with additional data-driven guidance, they are also capable of conditional generation (\cite{ding2025ccdmcontinuousconditionaldiffusion}). Current molecular generation methods, while advancing the field of drug discovery and materials science, face significant inefficiencies due to inherent limitations in their design and computational requirements. These challenges stem from the complexity of molecular structures, the high-dimensional search space, and the computational cost of existing models. As the size of the molecule increases, the search space grows exponentially, leading to longer generation times and higher resource consumption.

Variational autoencoders (VAEs) (\cite{kingma2023variationaldiffusionmodels}) have been widely used for molecular generation by encoding molecules into a continuous latent space and decoding them back to molecular structures. While VAEs have shown promise in molecular generation, they often struggle to generate chemically valid molecules due to inconsistencies between the continuous latent space and the discrete molecular structures. Generative Adversarial Networks (GANs) (\cite{goodfellow2016deep}) can generate chemically valid molecules by learning the underlying structure of chemical space, though they sometimes face challenges with mode collapse (\cite{inproceedings}; \cite{decao2022molganimplicitgenerativemodel}). These models often suffer from computational inefficiencies when scaling to larger molecules, especially in three-dimensional spaces. Although reinforcement learning (RL)-based methods (\cite{jaeger2024invitationdeepreinforcementlearning}) have shown promise in molecule generation (\cite{park2024molairmolecularreinforcementlearning}; \cite{mazuz2023molecule}; \cite{doi:10.1021/acs.jcim.1c01341}), several limitations hinder their effectiveness and generalization. 
Graph-based models generate molecular graphs sequentially, but as the graph size increases, generation times slow down and resource consumption rises (\cite{simonovsky2018graphvaegenerationsmallgraphs}; \cite{decao2022molganimplicitgenerativemodel}).  GeoDiff (\cite{xu2022geodiffgeometricdiffusionmodel}) achieves state-of-the-art performance in molecular conformer generation but faces significant scalability challenges. The iterative denoising process increases computational complexity as molecule size grows, while enforcing roto-translational invariance adds further memory overhead. Graph Diffusion Transformer (Graph DiT) (\cite{liu2024graphdiffusiontransformersmulticonditional}) enables multi-conditional molecular generation by integrating a graph-dependent noise model and a Transformer-based denoiser, achieving superior performance across various molecular properties. However, its scalability is limited due to the quadratic complexity of the Transformer architecture and the inefficiency of iterative denoising, especially for large and complex molecules.  

State-space models (SSMs) provide a robust framework for analyzing dynamical systems by representing the system's temporal dynamics through the evolution of latent states. These latent states govern the observed variables, enabling a structured approach to model complex, time-dependent processes (\cite{lin2024deeplearningbasedapproachesstate}). Latent space modeling is a fundamental technique used to capture the core features and underlying patterns of complex data by mapping them to low-dimensional representations. Structured state space sequence models (SSMs) (\cite{gu2022efficientlymodelinglongsequences}; \cite{nguyen2024alignment}; \cite{gu2024mambalineartimesequencemodeling}) have recently gained attention as an effective framework for sequence modeling. They can be viewed as bridging ideas from recurrent neural networks (RNNs) and convolutional neural networks (CNNs), while drawing on the foundations of classical state space models (\cite{10.1115/1.3662552}).

While most current molecule generation models are designed for small molecules, they fail to scale efficiently to large systems like polymers due to increased complexity, iterative bottlenecks, and gradient challenges during the iterative process (\cite{jin2020multiobjectivemoleculegenerationusing}; \cite{yang2023molecule}). 

\paragraph {Contributions.} In this paper, we propose a scalable and computationally efficient framework for molecule generation based on a Directly Denoising Diffusion Model (DDDM) by \cite{zhang2024directlydenoisingdiffusionmodels}. Our approach leverages the efficiency of few-step generation while preserving the quality and robustness of traditional diffusion techniques. By optimizing the denoising process and minimizing redundant computations, our framework achieves significant reductions in runtime and memory usage without sacrificing performance. The main focus of this work is to represent a step forward in addressing the computational challenges associated with molecular generation, enabling the scalable and efficient design of novel molecules. Our main contributions are as follows: 
\begin{itemize}
    \item We introduce a SE(3)-equivariant SSM-based diffusion framework, offering superior scalability and efficiency for molecule generation. Our approach aims at efficiently capturing long-range dependencies in molecular graphs, addressing the computational bottlenecks of traditional attention mechanisms. By integrating an input-dependent node selection mechanism derived from structured state-space models (SSMs), the model improves node-context reasoning in graphs, leading to more efficient and accurate conditional generation. 
    \item We establish theoretical guarantees on the efficiency of the diffusion reverse-process decomposition, showing that under uniform regularity conditions on the learned denoising maps, the reverse process can be reduced to a small number of well-structured deterministic subproblems.
    \item We evaluate the framework along both efficiency and quality dimensions, presenting results on inference speed, training FLOPs, and standard generation quality metrics. 
\end{itemize}
We report extensive benchmarking results on the GEOM dataset, covering both conventional molecular generation metrics (validity, novelty, diversity, QED) and 3D-specific evaluation measures such as COV–MAT, AMR, providing a comprehensive assessment of model quality and generalization. We conduct direct head-to-head comparisons with molecular diffusion baselines, including both 2D graph-based and 3D geometry-aware models, highlighting the efficiency and generative fidelity of our SE(3)-equivariant SSM framework. Further, our study provides a systematic investigation of SSM-based architectures for 3D molecular generation, revealing that such models can achieve favorable trade-offs between generative quality and computational efficiency during both training and inference. We construct GEOM-LongRange, a benchmark dataset derived from GEOM-MoleculeNet, containing molecules with more than 100 atoms. This dataset is designed to evaluate the model’s ability to capture long-range dependencies and generalize to larger, structurally complex molecular systems. 

\section{Background and Related Work}
Recent advances in molecule generation have leveraged deep learning techniques to address the complexity of designing novel chemical structures. Methods such as variational autoencoders (VAEs), generative adversarial networks (GANs), and reinforcement learning approaches (\cite{zhou2019moldqn}), have demonstrated promising results but face challenges related to efficiency, scalability, and chemical validity. More recently, diffusion models have emerged as a state-of-the-art approach for generating molecular structures due to their ability to model complex data distributions effectively (\cite{ho2020denoisingdiffusionprobabilisticmodels}, \cite{sohldickstein2015deepunsupervisedlearningusing}). However, these models often suffer from computational inefficiencies when scaling to larger molecules, especially in three-dimensional spaces. State-space models (SSMs) provide a robust framework for analyzing dynamical systems by representing the system’s temporal dynamics through the evolution of latent states. These latent states govern the observed variables, enabling a structured approach to model complex, time-dependent processes (\cite{lin2024deeplearningbasedapproachesstate}). Latent space modeling is a fundamental technique used to capture the core features and underlying patterns of complex data by mapping them to low-dimensional representations.

\textbf{Latent Space Diffusion Modeling}. In latent space diffusion models (\cite{pombala2025exploringmoleculegenerationusing}), data are mapped to a compressed latent representation where the diffusion process unfolds, improving efficiency and scalability. EDM-SyCo (\cite{ketata2024liftmoleculesmoleculargraph}) introduces a synthetic coordinate embedding framework that maps molecular graphs to Euclidean point clouds via synthetic conformer coordinates and learns the inverse mapping using an E(n)-Equivariant Graph Neural Network (EGNN). GeoLDM (\cite{hoogeboom2022equivariantdiffusionmoleculegeneration}) extends this idea to molecular geometry, combining autoencoders and diffusion models operating in latent space. CPDiffusion-SS (\cite{hu2024secondarystructureguidednovelprotein}) applies latent diffusion to protein sequence generation, leveraging secondary structural information for flexible and diverse outputs. Lalchand et al.     (\cite{lalchand2022recurrent}) proposed a variational sequential RNN-based architecture for de-novo drug design, mapping molecules to a continuous latent space for optimization and conditional generation. However, autoencoder and VAE-based methods often suffer from reconstruction errors, mode collapse, and limited diversity, while transformer-heavy architectures are computationally costly. In contrast, our model employs a scalable GraphGPS-based  (\cite{rampášek2023recipegeneralpowerfulscalable}) denoising network that integrates SE(3)-equivariant local message passing with efficient global attention, achieving accurate and efficient modeling of long-range molecular interactions.

\begin{figure}[h]
    \centering
    \includegraphics[width=0.98\textwidth]{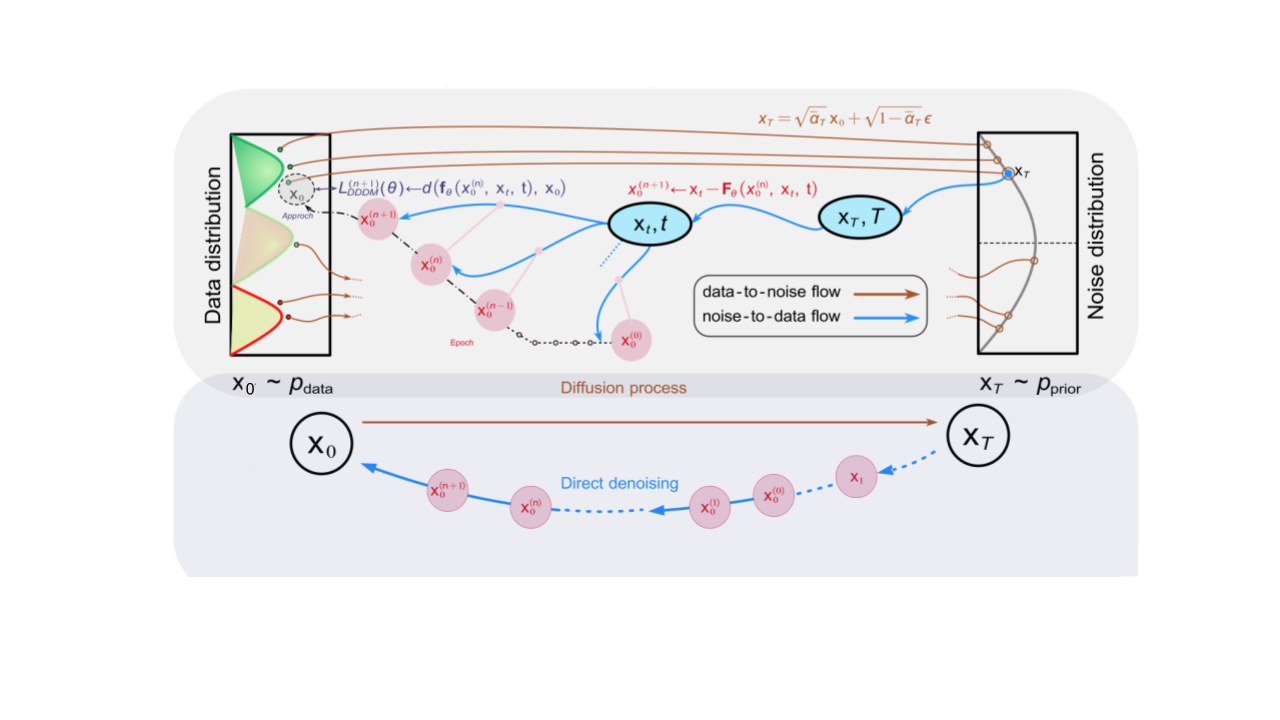}
    \vspace{-1cm}
    \caption{Overview of the diffusion generative process and reverse denoising dynamics. The top panel shows the forward noising trajectory from the data distribution $p_{\text{data}}$ to the noise prior $p_{\text{prior}}$, along with the corresponding 
reverse kernel approximation used during training. The bottom panel illustrates sampling via direct denoising from $x_T$ back to the clean data $x_0$.
}
\vspace{0.5cm}

    \label{fig:your_label}
\end{figure}

\section{Notations and Preliminaries}
\label{gen_inst}
\subsection{Diffusion Models}
\label{inst}
\paragraph{DDPMs}
Diffusion models (DDPMs) (\cite{ho2020denoisingdiffusionprobabilisticmodels}; \cite{sohldickstein2015deepunsupervisedlearningusing}) are generative models that learn data distributions via a forward process (adding noise) and a reverse process (removing noise). Let $\{\beta_i\}_{i=1}^T$ \text{ denote a sequence of positive noise scales such that }
$0 < \beta_1, \beta_2, \ldots, \beta_T < 1$. The forward process gradually adds Gaussian noise to the data $x_0$ over time steps $t$ in a Markov chain: $p(\mathbf{x}_t \mid \mathbf{x}_0)
= \mathcal{N}\!\left(
\mathbf{x}_t;\,
\sqrt{\bar{\alpha}_t}\,\mathbf{x}_0,\,
(1-\bar{\alpha}_t)\mathbf{I}
\right)$, 
where $\bar{\alpha}_t = \prod_{s=1}^{t}(1 - \beta_s)$ 
is the cumulative noise schedule and $x_{0} \sim p^{\ast} \propto \exp\!\left(-f^{\ast}\right)$ is a sample from the true data distribution and $f^*(x)$ is the corresponding energy function. We consider a discrete-time Markov chain ${x_0, x_1, \ldots, x_T}$. The perturbed data distribution is denoted as  $p_{\alpha}(\hat{\mathbf{x}}) := \int p^{\ast}(\mathbf{x})\, p_{\alpha}(\hat{\mathbf{x}} \mid \mathbf{x})\, d\mathbf{x}$.  A variational Markov chain in the reverse direction is parameterized as
$p_{\theta}(\mathbf{x}_{t-1} \mid \mathbf{x}_{t})
= \mathcal{N}\!\left(
\mathbf{x}_{t-1};
\frac{1}{\sqrt{1-\beta_t}}\bigl(\mathbf{x}_{t} + \beta_t\, s_{\theta}(\mathbf{x}_{t}, t)\bigr),\,
\beta_t \mathbf{I}
\right)$. Samples are produced via an ancestral sampling procedure, following
$\mathbf{x}_{t-1} = \frac{1}{\sqrt{1 - \beta_i}}
\left( \mathbf{x}_{t} + \beta_t \, s_{\theta^{\ast}}(\mathbf{x}_{t}, t) \right) + \sqrt{\beta_t} \, \mathbf{z}_t$. Here, $s_\theta(x_t, t)$ is the score function parameterized by $\theta$, which predicts the noise component in $x_t$. The optimal parameters $\theta^*$ are learned by minimizing the expected denoising error between the predicted and true noise over all timesteps.

\paragraph{Implicit Models} The induced joint distribution over latent variables is given by
$p(x_{1:T}\mid x_0) := p(x_T\mid x_0)\prod_{t=2}^{T} p(x_{t-1}\mid x_t, x_0)$.
The terminal marginal distribution is specified as
$p(x_T\mid x_0)=\mathcal{N}(\sqrt{\bar{\alpha}_T}\,x_0,(1-\bar{\alpha}_T)\mathbf{I})$, where
$\bar{\alpha}_t := \prod_{s=1}^{t}(1-\beta_s)$.
For all $t>1$, the backward kernels are defined as
$p(x_{t-1} \mid x_t, x_0)
=
\mathcal{N}\!\left(
\sqrt{\bar{\alpha}_{t-1}}\, x_0
+
\sqrt{1 - \bar{\alpha}_{t-1} - \beta_t}\,
\frac{x_t - \sqrt{\bar{\alpha}_{t}}\, x_0}{\sqrt{1 - \bar{\alpha}_{t}}},
\;\beta_t\,\mathbf{I}
\right)$.
The mean functions are constructed such that the induced marginals satisfy
$p(x_t\mid x_0)=\mathcal{N}(\sqrt{\bar{\alpha}_t}\,x_0,(1-\bar{\alpha}_t)\mathbf{I})$ for all $t$,
ensuring consistency across time steps.
Applying Bayes’ rule yields the induced forward transition
$p(x_t\mid x_{t-1},x_0)
=
\frac{p(x_{t-1}\mid x_t,x_0)\,p(x_t\mid x_0)}{p(x_{t-1}\mid x_0)}$. ~\cite{song2022denoisingdiffusionimplicitmodels} defined an oracle reverse-time inference process conditioned on the clean data sample
$x_0$, which serves as a theoretical reference distribution rather than a directly implementable
sampling algorithm.

\paragraph{Remark (Implicit Forward Process).}
\emph{
The forward transitions in implicit diffusion models do not correspond to the
Markovian noising process used in DDPMs. Instead, they arise from an oracle-based
inference construction and are therefore generally non-Markovian, with explicit
dependence on the clean sample \(x_0\).
A tractable generative procedure is obtained by replacing this oracle dependence
with a learned prediction \(\hat{x}_0 = f_\theta(x_t)\), yielding a reverse-time
process consistent with the DDIM parameterization.
In this setting, choosing \(\sigma_t^2 = \beta_t\) corresponds to the stochastic
formulation, while the limit \(\sigma_t \to 0\) recovers the deterministic
(implicit) case.
}

\textbf{Directly Denoising Diffusion}. From (\cite{song2020improvedtechniquestrainingscorebased}) reformulation of DDPMs as Stochastic Differential Equations (SDEs) to generalize the forward process: $d\mathbf{X}_t = -\frac{1}{2} \beta(t) \mathbf{X}_t \, dt + \sqrt{\beta(t)} \, d\mathbf{B}_t$, where $\mathbf{B}_t$ represents Brownian motion. The reverse-time VP SDE is: $d\mathbf{X}_t = \left[ -\frac{1}{2} \beta(t) (\mathbf{X}_t + \nabla \log p_t(\mathbf{X}_t)) \right] dt + \sqrt{\beta(t)} \, d\mathbf{B}_t$, where $\nabla \log p_t$ is estimated by a time-dependent score network $s_\theta(\mathbf{x}, t)$.  Directly Denoising Diffusion Models (DDDMs) (\cite{zhang2024directlydenoisingdiffusionmodels}) integrate DDPMs with the Probability Flow (PF) ODE framework, enabling faster denoising without complex solvers. The solution of the PF ODE is obtained by evaluating the integral expression $\mathbf{x}_0 = \mathbf{x}_T + \int_0^T -\frac{1}{2} \beta(t) [\mathbf{x}_t + \nabla_{\mathbf{x}_t} \log p_t(\mathbf{x}_t)] \, dt$, where $\mathbf{x}_T \sim \mathcal{N}(0, I)$. Directly Denoising Diffusion Models (DDDM) refine the estimate of the clean state $\mathbf{x}_0$ by leveraging the probability flow ODE. Specifically, the mapping $f(\mathbf{x}_0, \mathbf{x}_t, t) = \mathbf{x}_t - F(\mathbf{x}_0, \mathbf{x}_t, t)$ is defined, where $F$ involves an integral of the drift term parameterized by the noise schedule $\beta(t)$. A neural approximation $f_\theta(\mathbf{x}_0, \mathbf{x}_t, t) = \mathbf{x}_t - F_\theta(\mathbf{x}_0, \mathbf{x}_t, t)$ is trained such that $f_\theta \approx f$. 
\begin{equation}
\Theta := \arg\min_{\theta} \;
\mathbb{E}_{t \sim \mathcal{U}[1,T]} 
\Biggl[
  \mathbb{E}_{\mathbf{x}_0 \sim p_{\mathrm{data}}(\mathbf{x}_0)}
  \Biggl[
    \mathbb{E}_{\mathbf{x}_t \sim \mathcal{N}\Bigl(\sqrt{\bar{\alpha}_t}\mathbf{x}_0, (1-\bar{\alpha}_t)\mathbf{I}\Bigr)}
    \Bigl[ d\big(f_\theta(\mathbf{x}_0^{(n)}, \mathbf{x}_t, t), \mathbf{x}_0\big) \Bigr]
  \Biggr]
\Biggr]
\end{equation}
where, $d(\cdot, \cdot)$ is a suitable distance metric. We discuss choice of distance metrics in the Appendix Section \ref{distance_metrics}. We define $\Theta$ as the set of optimal neural parameters for the denoising network.
\subsection{Definitions}
The following section summarizes the notation and key preliminaries used in the remainder of this work.

\textbf{Definition 1 (Log-concavity).}
A function $f : \mathbb{R}^d \to \mathbb{R}_{+}$ is \emph{log-concave} if 
$\log f$ is concave. Equivalently, for all $x, y \in \mathbb{R}^d$ and 
$\lambda \in [0,1]$,
\[
    f(\lambda x + (1 - \lambda)y) 
    \;\ge\; 
    f(x)^{\lambda} f(y)^{1-\lambda}.
\]

\textbf{Definition 2 (Strong convexity).}
A twice-differentiable function $g : \mathbb{R}^d \to \mathbb{R}$ is 
\emph{$m$-strongly convex} if its Hessian satisfies
\[
    \nabla^{2} g(z) \succeq m I_d
    \qquad (m > 0).
\]
If $p(z) \propto \exp(-g(z))$, then $p$ is $m$-strongly log-concave.

\textbf{Definition 3 (Jacobian).}
A differentiable map $F : \mathbb{R}^d \to \mathbb{R}^d$ has Jacobian 
$J_F(z)=\nabla_z F(z)$ with entries 
$[J_F(z)]_{ij} = \partial F_i(z)/\partial z_j$.

\textbf{Definition 4 (Hessian).}
A twice-differentiable scalar function $g : \mathbb{R}^d \to \mathbb{R}$ has 
Hessian $\nabla^2 g(z) = [\partial^2 g(z)/(\partial z_i \partial z_j)]_{i,j=1}^d$.
For a vector-valued map $F$, the Hessian of its $i$th coordinate is 
$H_{F,i}(z) = \nabla^2 F_i(z)$.

\textbf{Definition 5 (Spectral and operator norms).}
For a matrix \(A \in \mathbb{R}^{d \times d}\), the spectral norm is
\(\|A\|_2 := \sup_{\|v\|=1} \|Av\|\).
For a second-derivative tensor \(D^2F(z)\), viewed as a symmetric bilinear map,
the operator norm is defined as
\(\|D^2F(z)\|_{\mathrm{op}} := \sup_{\|u\|=1,\;\|v\|=1} \|D^2F(z)[u,v]\|\).

\textbf{Definition 6 (Local Lipschitz constant).}
The local Lipschitz constant of $F$ at a point $z$ is 
$L(z) = \|J_F(z)\|_2$.

\textbf{Definition 7 (Residual).}
The residual in the reverse update is 
$r(z) = z - (x - F(z))$ 
with magnitude 
$R(z) = \|r(z)\|$.

\textbf{Definition 9 (Prékopa's theorem).}
By Prékopa's theorem, if the joint density \( q(z, x) \) is log-concave in the joint variables \((z, x)\), then the marginal density \( q(z) = \int q(z, x)\, dx \), obtained by integrating out \(x\), is also log-concave in \(z\).

\textbf{Definition 10 (Deterministic reverse update).}
An update of the form $x' = x - F(x')$ corresponds to the degenerate 
transition kernel 
$p(x'|x) = \delta(x' - (x - F(x')))$.
\begin{figure}[t]
\centering
\begin{minipage}[t]{0.52\linewidth}
\vspace{0pt}
\small
\textbf{Definition 8 (Reverse Transition Kernel (RTK) Inference).}
\cite{huang2024reversetransitionkernelflexible} proposes a flexible approach for accelerating diffusion inference by decomposing the reverse diffusion process into a small number of reverse transition kernel (RTK) sampling subproblems.
Under an appropriate decomposition, the number of such subproblems can be reduced to approximately
$\tilde{O}(1)$, independent of the diffusion discretization resolution, while ensuring that each RTK target
distribution remains strongly log-concave.

Strong log-concavity implies that each target density
$p(x) \propto \exp(-U(x))$ admits a strongly convex potential $U(x)$, yielding a unique global mode and
well-conditioned energy landscape. This enables fast mixing and provable convergence guarantees for
sampling algorithms such as Langevin dynamics and Metropolis--Hastings.
\end{minipage}
\hfill
\begin{minipage}[t]{0.44\linewidth}
\vspace{0pt}
\begin{algorithm}[H]
\caption{Reverse Diffusion via RTKs}
\label{alg:rtk_reverse_diffusion}
\small
\textbf{Setup:} Step size $\eta > 0$, horizon $T = K\eta$.

\textbf{for} $k = 0$ \textbf{to} $K-1$ \textbf{do}

\hspace{1em} Sample
$\hat{x}_{(k+1)\eta}
\sim \hat{p}_{(k+1)\eta \mid k\eta}
(\cdot \mid \hat{x}_{k\eta})$.

\hspace{1em} Assume marginal consistency:
$\hat{p}_{k\eta} \approx p_{(K-k)\eta}$.

\hspace{1em} Update marginal:
\[
\hat{p}_{(k+1)\eta}(z)
\approx
\int
p^{\leftarrow}_{(k+1)\eta \mid k\eta}(z \mid x)
\, \hat{p}_{k\eta}(x)\,dx .
\]

\textbf{end for}

\textbf{return} $\hat{X}_{K\eta}$
\end{algorithm}
\end{minipage}
\end{figure}

\subsection{Diffusion as Approximate Sampling of Transition Kernels}
Under this perspective, the denoising diffusion process is viewed as a sequence of sampling subproblems, where computational efficiency is governed by the trade-off between the number of subproblems and the difficulty of solving each one. DDPMs approximate each reverse transition using simple Gaussian kernels with $\mathcal{O}(1)$ per-step cost, but require $\tilde{\mathcal{O}}(d\,\varepsilon^{-2})$ such steps or subproblems to achieve $\varepsilon$-accuracy in total variation (TV) distance shown by (\cite{chen2023samplingeasylearningscore, benton2024nearlydlinearconvergencebounds}). 

In contrast, the Directly Denoising Diffusion Model (DDDM) can be interpreted within the Reverse Transition Kernel (RTK) framework as collapsing many small stochastic subproblems into a small number of deterministic kernel approximations. Rather than sampling from each intermediate reverse kernel, DDDM replaces stochastic transitions with a learned deterministic map that directly approximates the conditional mean of the RTK target distribution. From this viewpoint, DDDM performs approximate sampling by implicitly optimizing a transport map between successive noisy states, effectively solving fewer but harder subproblems. The RTK formulation clarifies that this deterministic denoising corresponds to the zero-noise limit of a valid reverse transition kernel, explaining how DDDM achieves efficient inference while remaining consistent with an underlying probabilistic model. 

\section{Problem Formulation}
In this section, we reformulate diffusion inference through the lens of reverse transition kernels (RTKs) and provide a principled connection between probability flow ODEs, Directly Denoising Diffusion Models (DDDMs), and kernel-based reverse diffusion. We begin by reviewing the probability flow ODE formulation, which yields a deterministic reverse-time dynamics governed by the score function. We show that an Euler discretization of this ODE naturally induces a sequence of reverse-time subproblems parameterized by a step size $\eta$. 

\subsection{Probability Flow ODE and Deterministic Reverse Diffusion}

Diffusion-based generative models are commonly formulated through a stochastic forward process that gradually perturbs data with noise, together with a reverse-time process that removes noise to recover samples from the data distribution. Following the stochastic differential equation (SDE) formulation of diffusion models introduced by (\citet{song2021scorebasedgenerativemodelingstochastic}), the variance-preserving (VP) forward diffusion process is given by
\begin{equation}
dX_t = -\frac{1}{2}\beta(t) X_t \, dt + \sqrt{\beta(t)} \, dB_t,
\end{equation}
where $\beta(t)$ is a time-dependent noise schedule and $B_t$ denotes standard Brownian motion. The corresponding reverse-time SDE takes the form
\begin{equation}
dX_t =
\left[
-\frac{1}{2}\beta(t)\left(X_t + \nabla_x \log p_t(X_t)\right)
\right] dt
+ \sqrt{\beta(t)} \, dB_t,
\end{equation}
where $\nabla_x \log p_t(x)$ denotes the score function of the marginal distribution at time $t$.

Song and Ermon (\citep{song2021scorebasedgenerativemodelingstochastic}) further showed that the reverse-time SDE admits an equivalent \emph{deterministic} formulation, known as the \emph{probability flow ordinary differential equation (PF ODE)}, which shares the same time-marginal distributions as the stochastic reverse process. The PF ODE is given by
\begin{equation}
d\overleftarrow{x}_t
=
-\frac{1}{2}\beta(t)
\left[
\overleftarrow{x}_t
-
\nabla_x \log p_t(\overleftarrow{x}_t)
\right] dt .
\label{eq:pf-ode}
\end{equation}
The marginal distributions of the forward and reverse processes are related through the identity
\begin{equation}
p_{T-t} = \overleftarrow{p}_t .
\end{equation}

In practice, the score function $\nabla_x \log p_t(x)$ is unknown and is approximated using a time-dependent neural network $s_\theta(x,t)$. Substituting this approximation into~\eqref{eq:pf-ode} yields a neural ordinary differential equation that governs deterministic reverse-time inference. Over a discrete time interval $t \in [k\eta, (k+1)\eta]$, the neural ODE can be written as
\begin{equation}
d\bar{x}_t
=
-\frac{1}{2}\beta(t)
\left[
\bar{x}_t
+
s_{\theta,\,T-k\eta}(\bar{x}_{k\eta})
\right] dt .
\label{eq:neural-ode}
\end{equation}

\subsection{Euler Discretization and Directly Denoising Diffusion Models}

Let $T$ denote the total diffusion horizon and let $\eta > 0$ be a step size such that $K = T / \eta$. Applying a forward Euler discretization to the probability flow ODE yields the approximation
\begin{equation}
\hat{x}_K
\approx
x_{k\eta}
-
\sum_{k=0}^{K-1}
\eta\, \beta(k\eta)
\left[
x_{k\eta}
-
\nabla_x \log p_{k\eta}(x_{k\eta})
\right].
\label{eq:euler-pf}
\end{equation}
Smaller step sizes $\eta$ lead to simpler local subproblems but require a larger number of steps, whereas larger step sizes reduce the number of steps at the cost of more challenging local approximations. 

Directly Denoising Diffusion Models (DDDMs) (\citep{zhang2024directlydenoisingdiffusionmodels}) are built on this discretized probability flow formulation. Instead of simulating the stochastic reverse-time SDE as in DDPMs, DDDMs perform deterministic denoising updates that directly approximate the solution of the PF ODE, thereby avoiding stochastic sampling or MCMC-based solvers during inference. At reverse-time step $k$, the DDDM update is given by
\begin{equation}
\hat{x}_{(k+1)\eta}
=
{x}_{k\eta}
-
F_{\theta, k\eta}(\hat{x}_{k\eta}),
\label{eq:dddm-update}
\end{equation}
where $F_{\theta, k\eta}$ is a neural approximation of the integrated drift term induced by the PF ODE and the noise schedule $\beta(t)$. Here, $\hat{x}_{k\eta}$ denotes the current estimate obtained from the previous reverse-time iteration, which serves as the input to the deterministic denoising update.

\subsection{Reverse Transition Kernel Perspective}

From the perspective of reverse transition kernels (RTKs), standard diffusion models such as DDPMs define stochastic reverse kernels that require sampling from conditional distributions at every step. In contrast, the deterministic update in Eq. \ref{eq:dddm-update} corresponds to a degenerate reverse transition kernel, which can be expressed as a Dirac measure:
\begin{equation}
\overleftarrow{p}_{(k+1)\eta \mid k\eta}(z \mid \hat{x}_{k\eta})
=
\delta\!\left(
z - \left({x}_{k\eta} - F_{\theta, k\eta}(\hat{x}_{k\eta})\right)
\right).
\label{eq:dirac-rtk}
\end{equation}
Thus, DDDM can be interpreted as a special case of the RTK framework in which each reverse transition kernel is deterministic and fully concentrated at the output of the denoising map.

\subsection{Proxy Reverse Kernels and Energy-Based Interpretation}

Although the exact reverse transition kernel induced by the PF ODE is intractable, the deterministic DDDM update admits a natural energy-based interpretation. We define the proxy energy function
\begin{equation}
g(z)
=
\frac{1}{2\sigma^2}
\left\|
z - \bigl(x_{k\eta} - F_{\theta, k\eta}(z)\bigr)
\right\|^2 ,
\label{eq:proxy-energy}
\end{equation}
which induces a tractable proxy reverse transition kernel
\begin{equation}
\hat{p}_{(k+1)\eta \mid k\eta}(z \mid \hat{x}_{k\eta})
\propto
\exp\bigl(-g(z)\bigr).
\label{eq:proxy-rtk}
\end{equation}
This formulation allows each deterministic DDDM update to be viewed as the solution of a well-structured reconstruction problem, rather than a stochastic sampling step.

\subsection{Implications for Reverse-Time Inference}
This reinterpretation of DDDM within the RTK framework underpins our theoretical analysis. Our goal in the following section is to show that, under mild regularity assumptions on the learned denoising maps $F_{\theta, k\eta}$, each reverse-time subproblem induced by Eq. \ref{eq:proxy-energy} admits a strongly log-concave target distribution. This property implies numerical stability, uniqueness of the reverse-time solution, and favorable optimization geometry. Crucially, strong log-concavity allows the reverse diffusion process to be decomposed into a small number of deterministic, well-conditioned subproblems, each of which can be solved efficiently without vanishing step sizes or long stochastic chains. As a result, inference in DDDM can be carried out using constant step sizes and $\tilde{O}(1)$ reverse-time subproblems, yielding substantially reduced computational complexity while preserving accuracy.

\begin{proposition}
\label{log_concavity}
(Log-Concavity of Reverse Subproblem(s)). DDDM reformulates the reverse process as a deterministic iterative refinement procedure, where each iteration corresponds to solving a strongly log-concave reconstruction objective. 

Let, $p(z) \propto \exp(-g(z))$ denote the target distribution of a reverse subproblem, where $g:\mathbb{R}^d \to \mathbb{R}$ is twice differentiable. We consider the proxy energy function $g(z) = \tfrac{1}{2\sigma^2}\|z - (x - F(z))\|^2$, where $F:\mathbb{R}^d \to \mathbb{R}^d$ is differentiable. Then $g$ admits the gradient $\nabla g(z) = \tfrac{1}{\sigma^2}(I + J_F(z))^\top r(z)$ and Hessian $\nabla^2 g(z) = \tfrac{1}{\sigma^2}\big((I + J_F(z))^\top (I + J_F(z)) + \sum_{i=1}^d r_i(z)\, H_{F,i}(z)\big)$, where $r_i(z) = z_i - x_i + F_i(z)$, $J_F(z)$ is the Jacobian of $F$, and $H_{F,i}(z)$ is the Hessian of its $i$-th output coordinate. To capture local geometric properties of F, we define the following quantities $L(z) := \|J_F(z)\|_2$, $R(z) := \|r(z)\|$, and $B_{\mathrm{sq}}(z) := \sum_{i=1}^d \|H_{F,i}(z)\|_{\mathrm{op}}^2$. Then, for all $v \in \mathbb{R}^d$, 
\begin{equation}
v^\top \nabla^2 g(z)\,v \geq \frac{1}{\sigma^2}\big((1 - L(z))^2 - R(z)B_{\mathrm{sq}}(z)\big)\|v\|^2.
\end{equation}

Hence, if $(1 - L(z))^2 > R(z)B_{\mathrm{sq}}(z)$, the energy function $g$ is $m$-strongly convex with $m = \tfrac{1}{\sigma^2}\big((1 - L(z))^2 - R(z)B_{\mathrm{sq}}(z)\big) > 0$, implying that the corresponding proxy kernel $\hat{p}(z \mid x) \propto \exp(-g(z))$ is $m$-strongly log-concave around $z$. Therefore, each reverse update in DDDM inherits strong log-concavity, ensuring convex and numerically stable behavior throughout the deterministic reverse trajectory.

where $\|\cdot\|_2$ denotes the spectral (operator) norm, and $L(z) = \|J_F(z)\|_2$ quantifies the local Lipschitz constant of $F$ around $z$.

\end{proposition}

\begin{proposition}
(Admissibility of constant step-size for DDDM). Consider the DDDM update at reverse-time step \(k\) with step-size \(\eta > 0\), given by \(\widehat{x}_{(k+1)\eta} = {x}_{k\eta} - F_{\theta,k\eta}(\widehat{x}_{k\eta})\), and the corresponding proxy energy \(g(z) = \tfrac{1}{2\sigma^2}\|z - (x_{k\eta} - F_{\theta,k\eta}(z))\|^2\), where \(F_{\theta,k\eta}:\mathbb{R}^d \to \mathbb{R}^d\) is twice continuously differentiable on the region of interest.
If,
\[
\exists\, \kappa \in [0,1),\; B \ge 0 \text{ such that } 
\|J_F(z)\|_2 \le \kappa \text{ and } 
\|D^2 F_{\theta,k\eta}(z)\|_{\mathrm{op}} \le B,\; 
\forall z \in \mathcal{Z}.
\]

where \(J_F(z)=\nabla_z F_{\theta,k\eta}(z)\) and \(\|D^2F\|_{\mathrm{op}}\) denotes an operator norm bound on the second derivative (tensor) action on unit vectors. Then under the condition, if,
\begin{equation}
\frac{(1 - \kappa)^2}{\sigma^2} > C_B
\label{eq:strong_convexity_condition}
\end{equation}
for a constant \(C_B\) that depends polynomially on \(B\) and \(d\), \(g\) is \(m\)-strongly convex on the region for some \(m>0\).
\end{proposition}
 Consequently the proxy density \(\propto e^{-g(z)}\) is sharply concentrated about a unique minimizer, and the transition kernel is (exactly or asymptotically) degenerate/Dirac. The admissibility of a constant step-size $\eta = \Theta(1)$ remains valid as long as the parameterized mappings $F_{\theta,k\eta}$ obey the aforementioned uniform smoothness and boundedness conditions for the given $\eta$. We relegate the proof of this proposition to the Appendix \ref{constant-step-size}.
\begin{lemma}
\label{reverse_marginals}
(Log-Concavity Preservation in Reverse Marginals).
Consider the reverse transition kernel $p^{\leftarrow}_{(k+1)\eta|k\eta}(z \mid x)$ and the current marginal $p^{\leftarrow}_{k\eta}(x)$. 
If both $p^{\leftarrow}_{(k+1)\eta|k\eta}(z \mid x)$ (as a function of $(z, x)$) and $p^{\leftarrow}_{k\eta}(x)$ are log-concave in their respective arguments, then by Prékopa’s theorem, the marginal distribution at step $k+1$, given by $\hat{p}_{(k+1)\eta}(z) \approx \int p^{\leftarrow}_{(k+1)\eta|k\eta}(z \mid x)\, p^{\leftarrow}_{k\eta}(x)\, dx$, is also log-concave in $z$.
\end{lemma}
\noindent
\textit{Remark.} By Prékopa’s theorem, marginalization of a log-concave function preserves log-concavity. Consequently, any lower-dimensional subproblem obtained through projection or conditioning in the reverse diffusion process remains log-concave.

\section{Experiments \& Methodology}
\subsection{Denoising Framework}
Our denoising framework is designed on the GraphGPS (\cite{rampášek2023recipegeneralpowerfulscalable}): an MPNN+Global Attention Hybrid. The GPS framework is built on three key components: (i) positional or structural encoding (SE/PE), (ii) a local message-passing mechanism, and (iii) a global attention mechanism. The processing modules of scalable GPS construct a computational graph that integrates message-passing graph neural networks (MPNNs) with Transformer-based global attention, using attention mechanisms with linear complexity $O(V)$ in the number of nodes. Our denoising framework replaces linear attention mechanisms in the GraphGPS framework with selective state space layers, enabling input-driven graph sparsification. Leveraging the inherent modularity of the GraphGPS framework, we replace the standard global attention layers with state-space model–based layers, including Mamba (\cite{gu2024mambalineartimesequencemodeling}), Mamba-2 (\cite{dao2024transformersssmsgeneralizedmodels}), Hydra, and Jamba. This modular design allows for seamless integration of alternative architectures, enabling systematic comparison across different state-space formulations within the same overall denoising pipeline. Unlike dense attention, where all nodes can attend to one another, most SSMs update each node only with information from preceding nodes, creating positional asymmetry in the available context. For SSMs with unidirectional scans, we employ two node-ordering schemes. In the importance-based scheme, nodes are ordered by heuristic scores (e.g., node degree or eigenvector centrality), with high-importance nodes placed later in the sequence to exploit richer contextual information. Alternatively, we follow from \cite{wang2024graphmambalongrangegraphsequence} and shuffle node ordering during training to preserve permutation invariance. In our framework, the node embeddings are augmented with structural or positional encodings (SE/PE). In addition, we add sinusoidal timestep embeddings into the inputs of our model to provide explicit temporal conditioning. 

\subsection{Experimental Setup}
\label{experimental_setup}
For the training and evaluation of our models, we utilize the GEOM dataset (\cite{Axelrod2020GEOMEM}), a collection of high-quality molecular conformations generated using metadynamics within the CREST software (\cite{C9CP06869D}). The node and edge chemical features, $f_a$ and $f_{ab}$, follow the construction described in (\cite{jing2023torsionaldiffusionmolecularconformer}). The node features include atom identity, atomic number, aromaticity, degree, hybridization, implicit valence, formal charge, ring membership, and ring size, yielding a $74$-dimensional feature vector for the GEOM\textsc{-drugs} subset. Edge features are represented as a $4$-dimensional one\hyp hot encoding of the bond type. We adopt an $84\%$/$10\%$/$6\%$ split for the training, validation, and test sets, respectively, and all models are trained on approximately $55{,}000$ molecules from the training split. We also construct a dataset from GEOM-MoleculeNet containing molecules with more than 100 atoms to assess the ability of our models to generalize to larger molecules and we name the dataset as GEOM-LongRange.

\subsection{Evaluation Metrics \& Baselines}
To evaluate the similarity between generated conformers and the reference conformers, we follow the train/validation/test splits mentioned in Section \ref{experimental_setup} and use two ensemble-based metrics: Average Minimum RMSD (AMR) and Coverage. Each metric is computed in terms of both Recall (R)—which quantifies how well the generated ensemble spans the diversity of the ground-truth conformers—and Precision (P)—which reflects how accurately the generated conformers match the reference structures.  In addition, we also report the standard GEOM-Drugs generation metrics, enabling direct comparison with established conformer-generation benchmarks. Among cheminformatics methods, we evaluate \textsc{RDKit} ETKDG (\cite{riniker2015better}),
the most widely used open-source conformer generation algorithm, as well as
\textsc{OMEGA} (\cite{article}), a commercial conformer
generation package that has been continuously developed and optimized over more than a
decade. Among machine learning--based approaches, we include \textsc{GeoMol} (\cite{ganea2021geomol}), \textsc{GeoDiff} (\cite{xu2022geodiffgeometricdiffusionmodel}) and torsional diffusion (\cite{jing2023torsionaldiffusionmolecularconformer}). Tables
\ref{conformer_metrics} and \ref{conformer_mterics_1} report the evaluation metrics for the generated conformer ensembles. Tables~\ref{tab:mixtral-llama} and~\ref{tab:geom-longrange} report performance on the GEOM-DRUGS and GEOM-LongRange datasets, respectively. Results on GEOM-QM9 are provided in the appendix.

\begin{table}[t]
\centering
\footnotesize
\caption{We evaluate the quality of the generated conformer ensembles on the \textsc{GEOM-Drugs} test set using 
\textit{Coverage (\%)} and \textit{Average Minimum RMSD (\AA)}. Following prior work, Coverage is computed 
with a tighter threshold of $\delta = 0.75$~\AA\ to provide clearer separation between high-performing methods.}
\label{conformer_metrics}
\resizebox{\textwidth}{!}{%
\begin{tabular}{lcccccccc}
\toprule
& \multicolumn{4}{c}{\textbf{Recall}} & \multicolumn{4}{c}{\textbf{Precision}} \\
\cmidrule(lr){2-5} \cmidrule(lr){6-9}
\textbf{Method} 
& \textbf{Cov $\uparrow$} & \textbf{AMR $\downarrow$} 
& \textbf{Cov $\uparrow$} & \textbf{AMR $\downarrow$}
& \textbf{Cov $\uparrow$} & \textbf{AMR $\downarrow$} 
& \textbf{Cov $\uparrow$} & \textbf{AMR $\downarrow$} \\
& \textbf{Mean} & \textbf{Mean} & \textbf{Med} & \textbf{Med}
& \textbf{Mean} & \textbf{Mean} & \textbf{Med} & \textbf{Med} \\
\midrule
RDKit ETKDG         & 38.4 & 1.058 & 28.6 & 1.002 & 40.9 & 0.995 & 30.8 & 0.895 \\
OMEGA               & 53.4 & 0.841 & 54.6 & 0.762 & 40.5 & 0.946 & 33.3 & 0.854 \\
GeoMol              & 44.6 & 0.875 & 41.7 & 0.834 & 43.0 & 0.928 & 36.4 & 0.841 \\
GeoDiff             & 41.4 & 0.945 & 37.8 & 0.903 & 44.5 & 1.136 & 41.5 & 1.090 \\
Torsional Diffusion & 72.7 & 0.582 & 80.0 & 0.565
                    & 55.2 & 0.778 & 56.9 & 0.729 \\
Our method (denoiser: EGNN + Hydra) & 89.7 & 0.232 & 92.0 & 0.723
                    & 74.6 & 0.456 & 78.9 & 0.344 \\                    
Our method (denoiser: EGNN + Jamba) & 82.7 & 0.466 & 87.8 & 0.665
                    & 69.9 & 0.275 & 74.9 & 0.534 \\                      
\bottomrule
\end{tabular}}
\end{table}

\begin{table}[t]
\centering
\caption{Median AMR and runtime (core-secs per conformer) of machine learning methods, 
evaluated on CPU for comparison with RDKit.}
\label{conformer_mterics_1}
\begin{tabular}{lccccc}
\toprule
\textbf{Method} & \textbf{Params (M)} & \textbf{Steps} & \textbf{AMR-R} & \textbf{AMR-P} & \textbf{Runtime} \\
\midrule
RDKit      & --   & --   & 1.002 & 0.895 & 0.10 \\
GeoMol     & 0.3   & --   & 0.834 & 0.841 & 0.18 \\
GeoDiff    & 1.6   & 5000 & 0.809 & 1.090 & 3.05 \\
\midrule
Torsional Diffusion   & 1.6 & 5  & 0.685 & 0.963 & 1.76 \\
Torsional Diffusion  & 1.6 & 10 & 0.580 & 0.791 & 2.82 \\
Torsional Diffusion  & 1.6 & 20 & 0.565 & 0.729 & 4.90 \\
\midrule
EGNN + Jamba  & 71.7 & 100  & 0.823 & 0.632 & 7.77 \\
EGNN + Jamba  & 71.7 & 200  & 0.765 & 0.591 & 15.54 \\
EGNN + Jamba  & 71.7 & 700  & 0.778 & 0.529 & 54.39 \\
\bottomrule
\end{tabular}
\end{table}
\section{Performance Analysis}
\begin{table*}[t]
\centering
\caption{Performance analysis of different architectures on GEOM-DRUGS. Higher is better for all metrics ($\uparrow$). All models are trained using the pseudo-Huber loss with parameter $c = 6.9 \times 10^{-5}$.}
\label{tab:mixtral-llama}
\footnotesize
\setlength{\tabcolsep}{3pt}
\renewcommand{\arraystretch}{1.1}
\resizebox{\textwidth}{!}{%
\begin{tabular}{lcccccccc}
\hline
\textbf{Model} &
\textbf{Train Params} &
\textbf{Novelty}~($\uparrow$) &
\textbf{Validity}~($\uparrow$) &
\textbf{Diversity}~($\uparrow$) &
\textbf{Atom.~Stab.}~($\uparrow$) &
\textbf{Mol.~Stab.}~($\uparrow$) &
\textbf{QED}~($\uparrow$) \\
\hline
EGNN + Transformer &
63.0M &
$85.4\pm0.14$\% &
$87.1\pm0.05$\% &
$89.5\pm0.08$\% &
-- &
-- &
$0.456\pm0.06$ \\

EGNN + Mamba-1 &
63.4M &
$83.6\pm0.09$\% &
$89.8\pm0.06$\% &
$92.1\pm0.06$\% &
-- &
-- &
$0.372\pm0.07$ \\

EGNN + Mamba-2 &
63.2M &
$90.8\pm0.02$\% &
$88.7\pm0.09$\% &
$92.6\pm0.07$\% &
-- &
-- &
$0.465\pm0.03$ \\

EGNN + Jamba &
71.7M &
$93.9\pm0.15$\% &
$95.4\pm0.02$\% &
$94.6\pm0.12$\% &
-- &
-- &
$0.637\pm0.05$ \\

EGNN + Hydra &
55.7M &
$94.5\pm0.07$\% &
$97.8\pm0.09$\% &
$94.4\pm0.05$\% &
-- &
-- &
$0.556\pm0.01$ \\
\hline
\end{tabular}
}
\end{table*}

Table~\ref{tab:mixtral-llama} shows that architectures with dense or bi-directional attention mechanisms, such as Transformers, Jamba, and Hydra, consistently achieve high performance across novelty, validity, diversity, and QED metrics. These models effectively capture long-range dependencies, allowing all nodes in the molecular graph to influence each other, which is crucial for accurate molecular generation. In contrast, the Mamba-based architectures (Mamba-1 and Mamba-2) lag behind on some metrics. This can be attributed to the recency bias (\cite{wang2025understandingmitigatingbottlenecksstate}) inherent in unidirectional SSMs: each node is updated sequentially based only on preceding node, so earlier nodes in the sequence have limited context. While node prioritization (See Appendix \ref{node_permutation_and_prioritization}) helps emphasize important nodes (placed later in the sequence), Mamba tends to forget nodes that appear earlier, resulting in suboptimal long-range interaction modeling.

\begin{table*}[t]
\centering
\caption{Performance analysis of different architectures on GEOM-LongRange. Higher is better for all metrics ($\uparrow$). All models are trained using the pseudo-Huber loss with parameter $c = 6.9 \times 10^{-5}$.}
\label{tab:geom-longrange}
\scriptsize
\setlength{\tabcolsep}{3pt}
\renewcommand{\arraystretch}{1.1}
\begin{tabular}{lcccccccc}
\hline
\textbf{Model} &
\textbf{Train Params} &
\textbf{Nov.}~($\uparrow$) &
\textbf{Val.}~($\uparrow$) &
\textbf{Div.}~($\uparrow$) &
\textbf{Atom.~Stab.}~($\uparrow$) &
\textbf{Mol.~Stab.}~($\uparrow$) &
\textbf{QED}~($\uparrow$) \\
\hline
EGNN + Transformer &
63.0M &
$94.4\pm0.09$\% &
$92.1\pm0.06$\% &
$89.5\pm0.07$\% &
-- &
-- &
$0.452\pm0.08$ \\

EGNN + Mamba-1 &
63.4M &
$95.6\pm0.03$\% &
$90.7\pm0.05$\% &
$92.1\pm0.07$\% &
-- &
-- &
$0.553\pm0.01$ \\

EGNN + Mamba-2 &
63.2M &
$86.8\pm0.02$\% &
$93.7\pm0.03$\% &
$92.6\pm0.07$\% &
-- &
-- &
$0.464\pm0.07$ \\

EGNN + Jamba &
71.7M &
$99.9\pm0.04$\% &
$95.4\pm0.02$\% &
$97.2\pm0.05$\% &
-- &
-- &
$0.525\pm0.01$ \\

EGNN + Hydra &
55.7M &
$92.5\pm0.09$\% &
$91.9\pm0.06$\% &
$94.4\pm0.03$\% &
-- &
-- &
$0.678\pm0.01$ \\
\hline
\end{tabular}
\end{table*}

In contrast, architectures such as Jamba overcome this limitation by combining the efficiency of SSMs with the global receptive field of attention. The attention pathway enables every node in the molecular graph to communicate with every other node, compensating for the directionality constraint of the underlying SSM block. This restores the ability to model symmetric, long-range structural dependencies that are essential for 3D molecular geometry. As a result, Jamba achieves substantially better performance on metrics that require global coherence and nonlocal interactions.

Hydra goes a step further by replacing semiseparable state mixing with quasiseparable matrices, which support intrinsic bidirectional information flow. Unlike unidirectional SSMs, Hydra’s quasiseparable formulation naturally mixes information from both earlier and later nodes in the sequence, eliminating recency bias at the architectural level. This richer bidirectional structure allows Hydra to capture long-range dependencies without needing explicit attention layers, and it does so with subquadratic efficiency. Because molecular graphs rely heavily on global, multi-hop context, this bidirectional mixing is particularly effective, explaining why Hydra often matches or surpasses attention-based models in structural accuracy despite being far more efficient.

\section{Denoising Architecture and Reverse-Time Inference}
\textbf{Rationale Behind the Denoising Framework}. The design of our denoising framework is motivated by the need to reconcile
\emph{theoretical stability in deterministic reverse diffusion} with
\emph{computational scalability on large molecular graphs}. As established in
Section~4, the Directly Denoising Diffusion Model (DDDM) reformulates the reverse
diffusion process as a deterministic reconstruction subproblem,
characterized by a strongly log-concave proxy energy under mild smoothness
and Lipschitz conditions. While this formulation guarantees numerical stability
and permits the use of large, constant step sizes in the reverse process, it
places stringent requirements on the denoising operator $F_{\theta,t}$, which
must simultaneously model fine-grained local structure and long-range global
dependencies at every reverse step.

Conventional graph denoising architectures based purely on local message passing
are limited in their ability to propagate global information efficiently, while
dense self-attention mechanisms incur quadratic complexity in the number of
nodes, rendering them impractical for large or high-resolution molecular graphs.
To address this trade-off, we adopt the GraphGPS framework, which explicitly
decouples \emph{local message passing} from \emph{global context aggregation}.
This separation allows each component to be optimized for its respective role in
the denoising dynamics and is particularly well suited to deterministic diffusion
models, where accurate global coordination is essential to prevent error
accumulation across reverse iterations.

\textbf{Step-size}. The admissibility of a constant step size $\eta = \Theta(1)$ in DDDM plays a crucial role in enabling efficient diffusion inference. When the parameterized mappings $F_{\theta,k\eta}$ satisfy the uniform smoothness and boundedness conditions, the associated proxy energy $g(z)$ becomes $m$-strongly convex, ensuring that the corresponding proxy density $\propto e^{-g(z)}$ is sharply concentrated around a unique minimizer. This strong convexity guarantees that each reverse-time update converges stably and effectively without requiring excessively small step sizes. Consequently, the denoising trajectory can be decomposed into only $\tilde{\mathcal{O}}(1)$ well-conditioned subproblems, as opposed to the $\tilde{\mathcal{O}}(1/\eta) = \tilde{\mathcal{O}}(1/\epsilon^2)$ subproblems required in standard DDPMs. This constant-step-size regime eliminates the need for finely discretized reverse diffusion schedules, substantially reducing the number of iterative updates while preserving numerical stability and accuracy. As a result, DDDM attains a provably faster inference process, enabling efficient sampling through a small number of strongly log-concave subproblems. In practice, depending on the smoothness of the learned field \(F_{\theta,k\eta}\) and the quality of its 
score approximation, each subproblem \(g_k(z)\) can typically be solved in a single deterministic update, or refined through a few inner iterations. DDIM achieves faster generation achieves faster generation by reducing the number of reverse transition kernels that must be sampled.Instead of solving $T$ Gaussian RTK subproblems (as in DDPM), DDIM constructs a sparse reverse trajectory and computes larger RTK jumps using closed-form, noise-free transitions. Because DDIM preserves the correct marginals $q(x_t \mid x_0)$, these coarse RTK steps remain consistent with the original diffusion model. Thus, 
DDIM is an example of RTK acceleration achieved through trajectory subsampling, analogous in spirit to the RTK--MALA (\cite{huang2024reversetransitionkernelflexible}) and RTK--ULD (\cite{huang2024reversetransitionkernelflexible}) approaches, which reduce the number of subproblems by employing stronger, more expressive kernels. 

\section{Conclusion and Future Work}
In theory, the entire reverse-time trajectory of DDDM from \(T \to 0\) can be discretized into \(K = T / \eta\) segments, where each step \(k\) defines a local deterministic subproblem governed by the proxy energy \(g_k(z)\). Thus, the reverse process can be interpreted as a sequence of \(K\) local energy minimization problems. Importantly, when the step size satisfies \(\eta = \Theta(1)\), the number of subproblems \(K\) remains small—often constant in order—while preserving stability and accuracy. Unlike standard DDPMs, which require $\mathcal{O}(1/\eta) = \mathcal{O}(1/\varepsilon^2)$ finely discretized stochastic subproblems to approximate the reverse diffusion SDE, DDDM replaces the stochastic kernel with the deterministic Probability Flow ODE. Each discrete step thus forms a single strongly log-concave subproblem whose minimizer defines the next iterate. When $F_{\theta,k\eta}$ satisfies uniform smoothness and boundedness conditions, these subproblems remain well-conditioned even under a constant step size $\eta = \Theta(1)$, allowing the entire reverse trajectory to be represented by only $\widetilde{\mathcal{O}}(1)$ deterministic updates. This constant-step-size regime eliminates the need for finely discretized reverse schedules, yielding a provably faster and more stable diffusion inference process. This work opens a promising direction for future research. While our analysis focuses on deterministic reverse-time dynamics under constant step sizes, extending the RTK formulation to hybrid stochastic–deterministic kernels could provide a principled trade-off between robustness and diversity, particularly for multimodal molecular distributions.

\bibliography{tmlr}
\bibliographystyle{tmlr}

\vspace{8cm} 
\appendix
\section{Appendix}
\subsection{More Related Work}
Recent advances in geometric learning and molecular modeling have increasingly focused on improving scalability and computational efficiency as tasks expand to larger molecules, protein complexes, and high-throughput generative pipelines. Several lines of work explore structural, spectral, and geometric strategies to achieve these goals. Dirac–Bianconi GNNs (\cite{nauck2024diracbianconi}) leverage the topological Dirac operator to model higher-order interactions beyond standard message passing, while commute-time-optimized graph (\cite{sterner2024commutetimeoptimisedgraphsgnns}) rewiring improves information flow by restructuring graphs based on commute-time priors. Global structural summarization methods such as G-Signatures (\cite{schäfl2023gsignaturesglobalgraphpropagation}) use randomized latent paths to capture whole-graph geometry efficiently. Other approaches enhance expressivity without sacrificing efficiency, including multivector neuron (\cite{liu2024multivectorneuronsbetterfaster}) models based on Clifford algebra and SE(3)-Hyena (\cite{moskalev2024se3hyenaoperatorscalableequivariant}), which introduces long-range equivariant convolutions with sub-quadratic complexity. In molecular modeling, lightweight generative frameworks such as AlphaFlow-Lit (\cite{li2024improvingalphaflowefficientprotein}) dramatically accelerate protein ensemble generation, and conformer-generation models like MCF (\cite{wang2024swallowingbitterpillsimplified}) show that scaling model capacity can improve 3D prediction without strong inductive biases. Collectively, these works highlight a growing emphasis on designing architectures that retain geometric fidelity while remaining computationally tractable for large-scale molecular systems.

A complementary line of research focuses on accelerating diffusion sampling to make generative models practical for large-scale, high-dimensional, or real-time molecular settings. Recent methods improve efficiency by reducing the number of sequential denoising steps or parallelizing them across hardware. Faster diffusion sampling (\cite{gupta2025faster}) algorithms inspired by the randomized midpoint method achieve sharply improved dimension dependence---reducing the complexity from $\tilde{O}(\sqrt{d})$ to $\tilde{O}(d^{5/12})$ in total variation distance---and provide the first provable guarantees for fully parallelizable diffusion sampling with a polylogarithmic runtime. \textsc{AsyncDiff} (\cite{chen2024asyncdiffparallelizingdiffusionmodels}) tackles a different bottleneck by exploiting model parallelism: it decomposes the noise predictor across devices and leverages the strong similarity of hidden states between adjacent diffusion steps, enabling asynchronous computation and significantly lowering latency without degrading generative fidelity. \textsc{ParaDiGMS} (\cite{shih2023parallelsamplingdiffusionmodels}) further accelerates inference by denoising several steps in parallel via Picard iterations, and is compatible with widely used accelerators such as DDIM and DPM-Solver, offering $2$--$4\times$ speedups while maintaining sample quality. Together, these approaches highlight the emerging trend of designing diffusion models that break sequential dependencies and enable scalable, parallel, and latency-efficient generation.

\subsection{Extended Discussion}
\textbf{GraphGPS Framework}. In our GraphGPS-based denoising architecture, all models share the same EGNN message-passing backbone, and differ only in the choice of global aggregation module. Although Transformers employ full dense attention, we find that the Jamba hybrid architecture consistently outperforms them across all molecular generation metrics. This is because dense attention, while theoretically capable of modeling global dependencies, often introduces noisy and overly uniform updates in large molecular graphs, leading to over-smoothing and reduced stability during iterative denoising steps. In contrast, Jamba integrates both an attention pathway and a state-space pathway, enabling complementary forms of information propagation: attention provides global, long-range interaction modeling, while the SSM branch supplies structured, smooth, and computationally stable updates. This hybridization yields stronger inductive biases, more robust propagation of molecule-level constraints, and improved preservation of chemically valid intermediate states. As a result, Jamba achieves higher validity, novelty, and QED than both Transformers and pure SSM variants, demonstrating that combining global attention with structured state-space propagation is more suited for diffusion-based 3D molecular generation.

\textbf{Unidirectional State-Space}. Unidirectional state-space models such as Mamba-1 and Mamba-2 inherently operate under a causal mixing structure, meaning that information flows only from earlier positions toward later ones. This induces a strong recency bias, where the model disproportionately attends to the most recent nodes in the input sequence while progressively forgetting earlier nodes. Although this temporal bias is suitable for language modeling, it becomes a clear limitation for molecular generation, where long-range atom–atom dependencies are symmetric and nonsequential. In molecules, chemically distant nodes often exert structural or electronic influence on one another, and these interactions do not obey a left-to-right ordering.

\textbf{DDDM}. For a deterministic DDDM setting, the denoising network must approximate a global transport map from noisy to clean molecular structures, often in a single step, with iterative refinement serving only as a lightweight correction mechanism. Under this regime, full Transformer attention performs suboptimally: despite its theoretical expressivity, dense attention tends to overmix node representations and amplify noise, making the one-step prediction unstable and causing iterative refinement to accumulate correlated errors. In contrast, Jamba’s hybrid architecture—combining global attention with structured state-space updates—offers a more stable inductive bias for deterministic transport. The SSM pathway produces smooth, low-variance updates that preserve molecular geometry, while the attention pathway selectively captures long-range chemical dependencies that are essential for accurate one-step reconstruction. This complementary structure significantly improves the stability and accuracy of both the initial one-step denoising and subsequent refinement passes, resulting in higher validity and structural fidelity compared to Transformer-based denoisers.

\textbf{DDIM}. Although DDIM employs the same Gaussian RTK approximation as DDPM, its key 
difference arises in the \emph{deterministic} limit $\sigma_t \to 0$. In this limit, 
the Gaussian kernel degenerates into a Dirac measure, yielding an implicit 
deterministic mapping
\begin{equation}
    x_{t-1} = \mu_{\mathrm{DDIM}}(x_t, x_0).
    \label{eq:ddim-update}
\end{equation}

\begin{algorithm}[H]
\caption{DDIM Sampling as a Reverse Transition Kernel (RTK)}
\label{alg:ddim-rtk}
\KwIn{
Noise predictor $\epsilon_\theta$; 
noise schedule $\{\bar{\alpha}_{k\eta}\}_{k=0}^K$; 
coarse grid $\tau=\{k_0=0<k_1<\cdots<k_S=K\}$ with \(S\ll T\)
}
\KwOut{Sample $\hat{x}_{K\eta}\sim p^*$}

\textbf{Initialization:}  
Sample $\hat{x}_{k_0\eta} \sim \mathcal{N}(0,I)$\;

\For{$i = 0$ \KwTo $S-1$}{
    \textbf{Clean sample estimation (RTK mean estimator):}
    \[
    \hat x_0^{(k)}:=f_\theta^{(k\eta)}(\hat x_{k\eta})=(\hat x_{k\eta}-\sqrt{1-\bar\alpha_{k\eta}}\,\epsilon_\theta^{(k\eta)}(\hat x_{k\eta}))/\sqrt{\bar\alpha_{k\eta}}
    \]

    \textbf{RTK transition (deterministic DDIM limit):}
    
    \[
    \hat{x}_{k_{i+1}\eta}
    =
    \sqrt{\bar{\alpha}_{k_{i+1}\eta}}\, \hat{x}_0^{(k_i)}
    +
    \sqrt{
        \frac{1-\bar{\alpha}_{k_{i+1}\eta}}
             {1-\bar{\alpha}_{k_i\eta}}
    }
    \left(
        \hat{x}_{k_i\eta}
        -
        \sqrt{\bar{\alpha}_{k_i\eta}}\, \hat{x}_0^{(k_i)}
    \right)
    \]
}

\Return $\hat{X}_{K\eta}$
\end{algorithm}

Since this mapping is closed-form and deterministic, it composes across 
time indices. Consequently, instead of solving $T$ RTK subproblems 
(as in DDPM), one may select a coarse sampling trajectory 
$\tau = \{k_0 < k_1 < \cdots < k_S\}$ with $S \ll T$ and apply 
the RTK map only along this subsequence:
\begin{equation}
    x_{k_{i+1}\eta}
    =
    \mu_{\mathrm{DDIM}}\!\left(
        x_{k_i\eta},\,
        x_0^{(k_i)}
    \right).
    \label{eq:ddim-coarse}
\end{equation}

Importantly, DDIM preserves the exact diffusion marginals 
$p(x_t \mid x_0)$, ensuring that these coarse RTK transitions remain 
consistent with the original diffusion process. Thus, DDIM accelerates 
generation by reducing the number of RTK subproblems while retaining 
model fidelity, mirroring the acceleration philosophy of the 
general RTK framework. 

\subsection{Proof of Proposition 1}
\begin{proof}
In the subsequent analysis, we omit the subscripts $\theta,k\eta$ for notational simplicity.
Let
\[
g(z)=\frac{1}{2\sigma^2}\|z-(x-F(z))\|^2=\frac{1}{2\sigma^2}\sum_{i=1}^d r_i(z)^2,\qquad
r_i(z)=z_i-x_i+F_i(z).
\]
Differentiating gives
\[
\partial_j g(z)=\frac{1}{\sigma^2}\sum_{i=1}^d r_i(z)\,\partial_j r_i(z),\qquad
\nabla g(z)=\frac{1}{\sigma^2}J_r(z)^\top r(z),
\]
where $(J_r)_{ij}=\partial_j r_i(z)$. A second differentiation and substitution
$\partial_j r_i(z)=\delta_{ij}+\partial_j F_i(z)$, $\partial_k\partial_j r_i(z)=\partial_k\partial_j F_i(z)$
yields the exact Hessian
\begin{equation}
\nabla^2 g(z)
= \frac{1}{\sigma^2}\Big((I + J_F(z))^\top (I + J_F(z)) + \sum_{i=1}^d r_i(z)\, H_{F,i}(z)\Big).
\label{eq:hessian-gz}
\end{equation}
where $J_F(z)=\nabla_z F(z)$ and $H_{F,i}(z)=\nabla^2 F_i(z)$ is the Hessian of the $i$-th output coordinate.

Consider an arbitrary direction $v\in\mathbb{R}^d$. Then
\begin{equation}
v^\top \nabla^2 g(z)\, v
= \frac{1}{\sigma^2}\Big(\| (I + J_F(z))v \|^2 
+ \sum_{i=1}^d r_i(z)\, v^\top H_{F,i}(z)\, v \Big).
\label{eq:quadratic-form-hessian}
\end{equation}
We bound the two terms separately.

\medskip\noindent\textbf{(I) Lower bound for the quadratic term.}
Using the reverse triangle inequality and the operator norm of $J_F$,
\[
\|(I+J_F)v\|=\|v+J_F v\|\ge \|v\|-\|J_F v\|\ge(1-\|J_F\|_{\mathrm{op}})\|v\|.
\]
Define $L(z):=\|J_F(z)\|_{\mathrm{op}}$; then
\[
\|(I+J_F)v\|^2 \ge (1-L(z))^2\|v\|^2.
\]

\medskip\noindent\textbf{(II) Upper bound for the correction (Hessian) term.}
Set $a_i := v^\top H_{F,i}(z)\,v$ (scalars). By the operator-norm bound on $H_{F,i}$,
\begin{equation}
|a_i| \le \|H_{F,i}(z)\|_{\mathrm{op}}\, \|v\|^2.
\label{eq:ai-bound}
\end{equation}
Thus
\begin{equation}
\sum_{i=1}^d a_i^2 
\le \|v\|^4 \sum_{i=1}^d \|H_{F,i}(z)\|_{\mathrm{op}}^2
= B_{\mathrm{sq}}(z)\, \|v\|^4.
\label{eq:ai-sum-bound}
\end{equation}
where we define $B_{\mathrm{sq}}(z):=\sum_{i=1}^d \|H_{F,i}(z)\|_{\mathrm{op}}^2$.
By Cauchy--Schwarz over the index $i$,
\begin{equation}
\Big|\sum_{i=1}^d r_i(z)\, a_i\Big|
\le \|r(z)\|_2\, \sqrt{\sum_{i=1}^d a_i^2}
\le R(z)\, B_{\mathrm{sq}}(z)\, \|v\|^2.
\label{eq:r-a-bound}
\end{equation}
where $R(z):=\|r(z)\|_2$.

\medskip\noindent\textbf{(III) Combine the two bounds.}
Inserting (I) and (II) into the expression for $v^\top\nabla^2 g(z)v$ gives
\[
v^\top\nabla^2 g(z)\,v
\ge \frac{1}{\sigma^2}\Big((1-L(z))^2 - R(z)\,B_{\mathrm{sq}}(z)\Big)\|v\|^2.
\]
Therefore if $(1-L(z))^2 > R(z)B_{\mathrm{sq}}(z)$ we obtain a positive lower bound; defining
\begin{equation}
m := \frac{1}{\sigma^2}\Big((1 - L(z))^2 - R(z) B_{\mathrm{sq}}(z)\Big) > 0.
\label{eq:strong-convexity-margin}
\end{equation}
yields $v^\top\nabla^2 g(z)v\ge m\|v\|^2$ for all $v$, i.e.\ $\nabla^2 g(z)\succeq m I_d$ and $g$ is $m$-strongly convex at $z$.
Consequently $p(z)\propto e^{-g(z)}$ is $m$-strongly log-concave in a neighborhood of $z$.

\medskip\noindent\textbf{Affine special case.}
If $F$ is affine, $F(z)=\Lambda z+b$ (so $J_F=\Lambda$ and $H_{F,i}=0$), then
\[
\nabla^2 g(z)=\frac{1}{\sigma^2}(I+\Lambda)^\top(I+\Lambda),
\qquad v^\top\nabla^2 g(z)v=\frac{1}{\sigma^2}\|(I+\Lambda)v\|^2\ge 0.
\]
Hence $g$ is convex. Moreover:
\begin{enumerate}
  \item[(a)] If $(I+\Lambda)$ is full-rank then $\|(I+\Lambda)v\|>0$ for all $v\neq0$, so $g$ is strictly (and hence strongly) convex.
  \item[(b)] If $(I + \Lambda)$ is not of full rank, then there exists a nonzero vector $v$ such that $(I + \Lambda)v = 0$. Consequently, the Hessian $\nabla^2 g(z)$ admits a zero eigenvalue corresponding to this direction, implying that $g$ is not strictly convex along $v$.

\end{enumerate}

\begin{corollary}[One-hidden-layer explicit bounds]
Consider the one-hidden-layer network
\begin{equation}
J_F(z) = W_2\, \mathrm{diag}\!\big(\phi'(u)\big)\, W_1, 
\qquad u = W_1 z + b_1.
\label{eq:jacobian-one-layer}
\end{equation}
such that the Jacobian operator norm admits the bound
\begin{equation}
\|J_F(z)\|_{\mathrm{op}}
\le \|W_2\|_{\mathrm{op}}\, \|W_1\|_{\mathrm{op}}\,
\max_j |\phi'(u_j)|.
\label{eq:jacobian-op-bound}
\end{equation}
The $i$-th component Hessian is given by
\begin{equation}
H_{F,i}(z)
= W_1^{\top}\, \mathrm{diag}\!\big(w_{2,i} \odot \phi''(u)\big)\, W_1.
\label{eq:hessian-component}
\end{equation}
and therefore satisfies
\begin{equation}
\|H_{F,i}(z)\|_{\mathrm{op}}
\le 
\|W_1\|_{\mathrm{op}}^2 \, \max_j |w_{2,i,j}\, \phi''(u_j)|.
\label{eq:hessian-op-bound}
\end{equation}
A uniform scalar curvature bound can thus be taken as
\begin{equation}
B 
= \|W_1\|_{\mathrm{op}}^2
\, \max_{i,j} |w_{2,i,j}|
\, \max_j |\phi''(u_j)|.
\label{eq:B-bound}
\end{equation}
Substituting into Eq.~\ref{eq:strong-convexity-margin} gives the explicit sufficient condition
\begin{equation}
\Big(1 
- \|W_2\|_{\mathrm{op}} \, \|W_1\|_{\mathrm{op}} \, \max_j |\phi'(u_j)|
\Big)^2 
> 
R \, \Big(
  \|W_1\|_{\mathrm{op}}^2 \, \max_{i,j} |w_{2,i,j}| \, \max_j |\phi''(u_j)|
\Big).
\label{eq:sufficient-condition}
\end{equation}
\footnote{For piecewise-linear nonlinearities such as ReLU or leaky ReLU, $\phi'' = 0$ almost everywhere, implying $B = 0$. The sufficient condition then simplifies to $\|J_F\|_{\mathrm{op}} < 1$, which is a global contraction condition ensuring convexity everywhere.}
\end{corollary}
\end{proof}
\subsection{Proof of Proposition 2}
\label{constant-step-size}
\begin{proof}
From the computation in the main text we have the exact Hessian
\[
\nabla^2 g(z)
=\frac{1}{\sigma^2}(I+J_F(z))^\top(I+J_F(z)) + R(z),
\qquad
R(z):=\frac{1}{\sigma^2}\sum_{i=1}^d r_i(z)\,H_{F,i}(z),
\]
where $J_F(z)=\nabla_z F_{\theta,k\eta}(z)$ and $H_{F,i}(z)=\nabla^2 F_i(z)$. 
The first matrix is a Gram matrix and therefore positive semidefinite:
\[
\frac{1}{\sigma^2}(I+J_F)^\top(I+J_F)\succeq 0.
\]

By assumption $\|J_F(z)\|_2\le\kappa<1$.  For any unit vector $v$ the reverse triangle inequality yields
\[
\|(I+J_F)v\|=\|v+J_F v\|\ge\big|\|v\|-\|J_F v\|\big|\ge 1-\|J_F\|_2\ge 1-\kappa,
\]
hence the smallest singular value of \(I+J_F\) satisfies \(\sigma_{\min}(I+J_F)\ge 1-\kappa\).

Using the definition of singular values of a matrix A (spectral inequality from SVD), 
\[
A^\top A \succeq \sigma_{\min}(A)^2 I.
\]
we obtain the spectral inequality
\[
(I+J_F)^\top(I+J_F)\succeq (1-\kappa)^2 I,
\]
Finally, scaling by  $\frac{1}{\sigma^{2}}$ 
\[
\frac{1}{\sigma^2}(I+J_F)^\top(I+J_F)\succeq \frac{(1-\kappa)^2}{\sigma^2} I.
\]

We now establish an operator-norm bound for the remainder term \(R(z)\). We assume that, over the region of interest, the following uniform bounds hold.
\[
\|H_{F,i}(z)\|_{\mathrm{op}}\le B\qquad\text{for }i=1,\dots,d,
\qquad\text{and}\qquad \|r(z)\|_2\le R_{\max}.
\]
Then, using triangle inequality and Cauchy--Schwarz,
\[
\big\| \sum_{i=1}^d r_i(z)\,H_{F,i}(z)\big\|_{\mathrm{op}}
\le \sum_{i=1}^d |r_i(z)|\,\|H_{F,i}(z)\|_{\mathrm{op}}
\le \|r(z)\|_2\sqrt{\sum_{i=1}^d\|H_{F,i}(z)\|_{\mathrm{op}}^2}
\le R_{\max}\sqrt{d}\,B.
\]
Including the factor \(1/\sigma^2\) from the definition of \(R(z)\) gives
\[
\|R(z)\|_{\mathrm{op}}\le \frac{R_{\max}\sqrt{d}\,B}{\sigma^2}.
\]
Hence, we can define
\[
C_B \;=\; \frac{R_{\max}\sqrt{d}\,B}{\sigma^2},
\]
such that \( \|R(z)\|_{\mathrm{op}}\le C_B\) uniformly on the region.
Combining the two bounds yields the matrix inequality
\[
\nabla^2 g(z)
\succeq \left(\frac{(1-\kappa)^2}{\sigma^2}-C_B\right) I.
\]
Therefore, if
\[
\frac{(1-\kappa)^2}{\sigma^2}>C_B,
\]
then \(\nabla^2 g(z)\) is uniformly positive definite on the region and \(g\) is \(m\)-strongly convex with modulus
\[
m \;=\; \frac{(1-\kappa)^2}{\sigma^2}-C_B \;>\;0.
\]
\end{proof}

\subsection{Node Prioritization and Permutation Strategy}
\label{node_permutation_and_prioritization}
Unlike dense attention mechanisms, where all nodes attend to one another, the Mamba architecture performs sequential updates where each node is conditioned only on the hidden states of preceding nodes. For an input sequence of \(n\) nodes, the last node (\(n - 1\)) receives the most contextual information, having access to all prior nodes \(0\) through \(n - 2\), while earlier nodes are limited to progressively less context. This inherent asymmetry allows the Graph-Mamba-based denoising framework to implicitly emphasize more important nodes—those positioned later in the sequence, thereby enabling more effective sparsification. To leverage this property, we implement the node prioritization strategy from (\cite{wang2024graphmambalongrangegraphsequence}) that orders nodes based on heuristic importance scores, such as \textit{node degree} or \textit{eigenvector centrality}. Given a graph \(\mathcal{G} = (\mathcal{V}, \mathcal{E})\), with \(\mathcal{V} = \{v_1, \dots, v_n\}\) and corresponding node embeddings \(x_i \in \mathbb{R}^d\), we compute a scalar score \(s_i\) for each node and define a permutation \(\pi\) such that \(s_{\pi(1)} \leq s_{\pi(2)} \leq \dots \leq s_{\pi(n)}\). The permuted sequence \((x_{\pi(1)}, \dots, x_{\pi(n)})\) is then passed through the Mamba module of the denoising model, which performs a \textit{unidirectional} state-space scan: \(h_t = f(x_{\pi(1)}, \dots, x_{\pi(t)})\), where each node's hidden state is updated based only on previous elements in the sequence. To maintain permutation invariance, nodes with identical importance scores are randomly shuffled during training. 

\subsection{Positional Encodings}
We follow the modular framework of GRAPHGPS (\cite{rampášek2023recipegeneralpowerfulscalable}) to incorporate the structural/positional encodings to the inputs of our denoising model. We use the graph-based Laplacian positional encodings. If \( L \) represent the Laplacian matrix of a given graph \( G = (V, E) \), as a symmetric and positive semidefinite (PSD) matrix, \( L \) can be decomposed using its eigenvalues and eigenvectors as:
\begin{equation}
L = \sum_i \lambda_i u_i u_i^T
\end{equation}
where \( \lambda_i \) denotes the eigenvalues and \( u_i \) corresponds to the associated eigenvectors. In the context of a unified framework for graph neural network (GNN) positional encodings, the normalized graph Laplacian is defined as:
\begin{equation}
L = I - D^{- \frac{1}{2}} A D^{- \frac{1}{2}} = U^T \Lambda U
\end{equation}
Here, each row of \( U \) represents a corresponding eigenvector of the graph, while \( \Lambda \) is a diagonal matrix containing the eigenvalues. Based on this formulation, the positional encoding for a node \( k \) can be represented as outlined in \cite{grötschla2024benchmarkingpositionalencodingsgnns}:
\begin{equation}
X_k^{PE} = f(U_{k,:}, \Lambda, \Theta, \{ \cdot \})
\end{equation}
In the above expression, \( U_{k,:} \) denotes the \( k \)-th row of \( U \), \( \Lambda \) holds the eigenvalues, and \( \Theta \) represents the parameters controlling the linear or non-linear transformations applied to these matrices. The term \( \{ \cdot \} \) refers to any additional parameters specific to different encoding approaches.
The transformer block employs sinusoidal positional encodings to capture token order, since it does not use recurrence or convolution. 
These encodings are added to token embeddings and are defined as
\begin{equation}
PE(pos, 2i)   = \sin\!\left(\tfrac{pos}{10000^{2i/d_{\text{model}}}}\right), 
\quad 
PE(pos, 2i+1) = \cos\!\left(\tfrac{pos}{10000^{2i/d_{\text{model}}}}\right).
\label{eq:positional_encoding}
\end{equation}
where $pos$ is the position index and $i$ the embedding dimension. In our implementation, we follow ~\cite{vaswani2023attentionneed} and apply sinusoidal positional encoding by constructing a positional vector of the same dimension as the input embedding, i.e., $|PE(i)| = |x_i| = d_e$, and then adding it element-wise such that $x_i^{p} = x_i + PE(i)$.

\subsection{Implementation of PE-Equivariance and PE-Stability}
To ensure the theoretical guarantees of permutation equivariance and stability (Definitions~3.1--3.3 from \cite{wang2022equivariantstablepositionalencoding}), 
we employ the \textbf{EquivStableLapPE} node encoder, which computes positional embeddings based on 
the eigen-decomposition of the graph Laplacian. 
Specifically, for a graph \( G = (A, X) \), let \( L = D - A \) denote the normalized Laplacian matrix, 
with eigenvalues \( \Lambda = \mathrm{diag}(\lambda_1, \ldots, \lambda_k) \) 
and corresponding eigenvectors \( U = [u_1, \ldots, u_k] \). 
The Laplacian eigenvectors \( U \) serve as orthogonal positional features that transform according 
to the orthogonal group \( O(k) \) under node permutations (see Definition~2.5) by \cite{wang2022equivariantstablepositionalencoding}. 

In our implementation, the \textbf{EquivStableLapPE} module takes the precomputed eigenvectors \( U \) 
(stored as \texttt{batch.EigVecs}) and applies a normalization layer (BatchNorm or identity) 
followed by a learnable linear projection:
\begin{equation}
Z = \mathrm{Linear}(\mathrm{Norm}(U)),
\end{equation}
where \( Z \in \mathbb{R}^{n \times d} \) represents the positional embeddings of all \( n \) nodes. 
This formulation preserves \textit{permutation equivariance}, since permuting the node order corresponds 
to left-multiplying \( U \) by a permutation matrix \( P \), i.e., \( PU \), 
which induces the same permutation on the encoded features \( Z \).

Moreover, stability is achieved by ensuring that small perturbations in the graph Laplacian 
(e.g., minor edge rewiring) lead to bounded changes in the positional embeddings due to the continuity 
of the eigenbasis and the Lipschitz continuity of the linear encoder. 
This satisfies the PE-stability criterion in Eq.~(1) (\cite{wang2022equivariantstablepositionalencoding}):
\begin{equation}
\| X^{(1)} - P^* X^{(2)} \|_F + \eta(Z^{(1)}, P^* Z^{(2)}) \le C \, d(G^{(1)}, G^{(2)}).
\end{equation}

\subsection{Expressivity and Scalability}
Scaling Transformers to long sequences has been a persistent challenge due to the quadratic runtime and memory cost of self-attention. Although recent LMs such as GPT-4 (32K) (\cite{openai2024gpt4technicalreport}), MPT (\cite{mosaicml2023mpt7b}), and Claude (100K) have extended context lengths, they rely heavily on efficiency optimizations like FlashAttention, which exploits GPU memory hierarchy to achieve near-linear attention and enables extremely fast training (e.g., FlashAttention-2 (\cite{dao2023flashattention2fasterattentionbetter}) reaches up to 225 trillion operations per second on a single A100).

For graph data, “context’’ corresponds to tokenized representations of nodes, edges, or subgraphs. While longer sequences should, in principle, provide richer structural information, Transformer-based graph models (including GTs) inherit the quadratic bottleneck and often fail to exploit long contexts effectively; performance can even stagnate or degrade as sequence length grows (\cite{shi2023largelanguagemodelseasily}).

SSMs offer a scalable alternative. Its selective state-space mechanism filters out irrelevant information and supports dynamic state resets, preventing the accumulation of redundant context. As a result, SSMs scale efficiently with sequence length and exhibits monotonically improving performance on longer sequences (\cite{gu2024mambalineartimesequencemodeling}). When graphs are serialized into long token sequences, SSMs can leverage both local and global structure more effectively (\cite{frasca2022understandingextendingsubgraphgnns}), making it a compelling replacement for Transformer architectures in graph representation learning.
\subsection{Architecture Details}
In this section, we discuss the architectures of our message passing and SSM module in detail. Modern architectures, like the Transformer, consist of two key elements or components that is: a sequence mixer (Multi-Head Attention) and a channel mixer (Feed-Forward Network). Most sequence mixers can be expressed as matrix multiplications of the form \( Y = MX \). This is often defined as the Matrix Mixer framework which includes sequence based models such as attention, state-space models etc. As discussed in \cite{hydra_matrix_mixer}, the key insight lies in the structure of the mixer matrix \(M\). Building on this foundation, \cite{hwang2024hydrabidirectionalstatespace} proposed a systematic methodology for designing new mixers or architectures. The major bottleneck lies the matrix multiplication in \(M\) which might incur quadratic costs. Self-Attention offers strong flexibility and effectiveness for sequence mixing, but comes at a high computational cost. Most modern sequence-based architectures rely on two core principles: data-dependent matrix mixing and scalability, allowing sequence mixers to generalize beyond their training lengths.

\begin{table}[h]
\caption{Configuration of the Mamba \& Mamba-2-based architectures} 
\label{tab:egnn-models}
\begin{center}
\scriptsize
\renewcommand{\arraystretch}{1.3}

\resizebox{0.7\textwidth}{!}{
\begin{tabular}{lccccc}
\hline
\textbf{Model} & \textbf{$L$} & $\mathbf{d_{\text{state}}}$ & $\mathbf{d_{\text{conv}}}$ & $\mathbf{d_{\text{model}}}$ & \textbf{Params} \\
\hline
EGNN + Transformer & 8 & --   & --   & 512 & 63.0M \\
EGNN + Mamba       & 6 & 512  & 256  & 512 & 63.4M \\
EGNN + Mamba-2     & 6 & 512  & 256  & 512 & 63.2M \\
EGNN + Mamba       & 6 & 256  & 128  & 256 & 53.0M \\
EGNN + Mamba-2     & 6 & 256  & 128  & 256 & 50.7M \\
EGNN + Mamba-2     & 6 & 128  & 64   & 128 & 47.5M \\
EGNN + Mamba       & 8 & 512  & 256  & 512 & 82.4M \\
EGNN + Mamba-2     & 8 & 512  & 256  & 512 & 83.2M \\
EGNN + Mamba       & 8 & 256  & 128  & 256 & 67.6M \\
EGNN + Mamba-2     & 8 & 256  & 128  & 256 & 67.5M \\
EGNN + Mamba       & 10 & 512  & 256 & 512 & 105.3M \\
EGNN + Mamba-2     & 10 & 512  & 256 & 512 & 105.5M \\
EGNN + Mamba       & 10 & 128  & 64  & 128 & 79.2M \\
EGNN + Mamba-2     & 10 & 128  & 64  & 128 & 80.1M \\
\hline
\end{tabular}
}
\end{center}
\end{table}

\begin{table}[h]
\caption{Configuration of the Jamba-based architectures}
\label{tab:jamba-configs}
\begin{center}
\setlength{\tabcolsep}{5pt} 
\renewcommand{\arraystretch}{1.1}
\scalebox{0.85}{ 
\begin{tabular}{lcccccccc}
\hline
\textbf{Model} & \textbf{$L$} & $\mathbf{d_{\text{state}}}$ & $\mathbf{d_{\text{conv}}}$ &
$\mathbf{d_{\text{model}}}$ & \textbf{Params} &
\textbf{\# Experts} & \textbf{Top-$K$} & \textbf{Active Params} \\
\hline
EGNN + Jamba & 8  & 512 & 256 & 512 & 2.15B   & 16 & 4 & 537.5M \\
EGNN + Jamba & 6  & 256 & 128 & 256 & 286.9M  & 8  & 2 & 71.7M \\
EGNN + Jamba & 10 & 512 & 256 & 512 & 1.68B   & 8  & 2 & 420M \\
EGNN + Jamba & 10 & 256 & 128 & 256 & 478M    & 8  & 2 & 119.5M \\
EGNN + Jamba & 8  & 128 & 64  & 128 & 144.5M  & 16 & 4 & 36.1M \\
\hline
\end{tabular}}
\end{center}
\end{table}

\begin{table}[h]
\caption{Configuration of the architectures for our denoising method} 
\label{tab:egnn-models}
\begin{center}
\scriptsize
\renewcommand{\arraystretch}{1.3}

\resizebox{0.75\textwidth}{!}{
\begin{tabular}{lcccccc}
\hline
\textbf{Model} & \textbf{$L$} & $\mathbf{d_{\text{state}}}$ & $\mathbf{d_{\text{conv}}}$ & $\mathbf{d_{\text{model}}}$ & \textbf{\# Heads} & \textbf{Params} \\
\hline
EGNN + Transformer & 8 & --   & --   & 512 & 8  & 63.0M \\
EGNN + Mamba       & 6 & 512  & 256  & 512 & 8  & 63.4M \\
EGNN + Mamba-2     & 6 & 512  & 256  & 512 & 8  & 63.2M \\
EGNN + Hydra       & 6 & 256  & 128  & 256 & 4  & 55.7M \\
EGNN + Jamba       & 6 & 256  & 128  & 256 & 4  & 286.9M \\
\hline
\end{tabular}
}
\end{center}
\end{table}

\subsubsection{Implementations for a Single 80GB GPU}
\textbf{Transformer}. Our Transformer block follows the encoder–decoder architecture of Vaswani et al. (\cite{vaswani2023attentionneed}). The model is composed of a stack of layers, where each layer contains a multi-head self-attention mechanism followed by a position-wise feed-forward network. Residual connections are employed around each of these sub-layers, and layer normalization is applied to the outputs. We set the channel dimension for our transformer block to 512.

\textbf{Mamba}. Our configuration employed a Mamba block with a state size of 256, model dimension of 256, and 4 attention heads. Our architecture follows Gu and Dao (\cite{gu2024mambalineartimesequencemodeling}) in omitting explicit positional encodings, while using RMSNorm (\cite{zhang2019rootmeansquarelayer}) for normalization. Mamba extends structured state space models by allowing parameters to vary as functions of the input, enabling content-based reasoning over discrete modalities. This selective mechanism allows the model to adaptively propagate or forget information depending on the token context. Mamba incorporates a hardware-optimized recurrent implementation that achieves linear scaling in sequence length and up to 5× faster inference than Transformers. The model dimension is expanded by an expansion factor, which we set to 2 for all our Mamba blocks. 

\textbf{Mamba-2}. For Mamba-2, we adopt the same overall architecture as in Mamba, with each layer substituted by the revised Mamba-2 block from Dao and Gu (\cite{dao2024transformersssmsgeneralizedmodels}).  We set the internal Mamba-2 state dimension to 128 and expansion factor 2. We set $d_{conv}$ as the dimensionality of the internal convolutional layer in each block to 64. 

\textbf{Hydra}. Hydra builds on the conceptual foundation of the Matrix Mixer Sequence Models. Hwang et al. (\cite{hwang2024hydrabidirectionalstatespace}) introduced Hydra which uses a quasiseparable matrix mixer framework. Quasiseparable matrices generalize both the low-rank matrix mixers of linear attention and the semiseparable matrices of state space models, used as bidirectional extension of the semiseparable matrices. Hydra restricts \(M\) to being a structured matrix which is essential for subquadratic matrix multiplication algorithms. The state dimension of the hydra module is 256 and $d_{conv}$ is  set to 128 and expansion factor 2.

\begin{table}[h]
\caption{Configuration of the Hydra-based architectures} \label{tab:egnn-models}
\begin{center}
\normalsize  
\renewcommand{\arraystretch}{1.25}  
\begin{tabular}{lccccc}
\hline
\textbf{Model} & \textbf{$L$} & $\mathbf{d_{\text{state}}}$ & $\mathbf{d_{\text{conv}}}$ & $\mathbf{d_{\text{model}}}$ & \textbf{Params} \\
\hline
EGNN + Hydra & 8  & 512 & 256 & 512 & 82.2M \\
EGNN + Hydra & 6  & 256 & 128 & 256 & 55.7M \\
EGNN + Hydra & 6  & 128 & 64  & 128 & 58.9M \\
EGNN + Hydra & 10 & 256 & 128 & 256 & 83.4M \\
EGNN + Hydra & 8  & 128 & 64  & 128 & 63.0M \\
\hline
\end{tabular}
\end{center}
\end{table}
\textbf{Jamba}. Jamba is a novel Attention-SSM hybrid architecture, which combines Transformer layers with the Mamba layers at a certain ratio along with the mixture-of-experts (MoE) (\cite{chen2022understandingmixtureexpertsdeep}) module. To improve model capacity without proportionally increasing computational cost, Jamba incorporates Mixture-of-Experts (MoE) layers in place of MLP blocks. Specifically, MoE is applied in alternating layers, with 8 experts per layer and the top-2 experts selected for each token, following (\cite{lieber2024jambahybridtransformermambalanguage}). MoE block is used every 2 layers instead of MLP. Also, following \cite{lieber2024jambahybridtransformermambalanguage}, each Jamba block used in our architecture is a combination of mamba and attention layers with the ratio of attention-to-Mamba layers set to 1:6. Each layer in the block is either a Mamba or an Attention layer, followed by a multi-layer perceptron (MLP). Jamba stabilizes large-scale training with RMSNorm in Mamba layers, removing the need for positional embeddings such as RoPE \footnote{ All experiments are conducted using the Jamba variant without explicit positional encodings, as the Mamba layers implicitly capture positional information.}. The architecture uses standard components, including (Group Query Attention) GQA (\cite{chen2024optimisedgroupedqueryattentionmechanism}), SwiGLU activation function (\cite{shazeer2020gluvariantsimprovetransformer}), and MoE load balancing.

\subsubsection{Message Passing Module}
Our message passing module is designed to maintain equivariance to both rotations and translations of the node
coordinates $\mathbf{x}_i$, while also preserving equivariance to permutations over the set of nodes
$V$, similar to standard GNNs. At the core of this architecture lies the Equivariant Graph
Convolutional Layer (EGCL) from \cite{satorras2022enequivariantgraphneural} , which operates on the node feature embeddings
$\mathbf{h}^l = \{\mathbf{h}_0^l, \ldots, \mathbf{h}_{M-1}^l\}$, the coordinate embeddings
$\mathbf{x}^l = \{\mathbf{x}_0^l, \ldots, \mathbf{x}_{M-1}^l\}$, and the edge information
$E = \{e_{ij}\}$. The EGCL updates both the node features and the corresponding coordinates using
the following operations:
\begin{align}
\mathbf{m}_{ij} &= \phi_e\!\left(
\mathbf{h}_i^l,\,
\mathbf{h}_j^l,\,
\lVert \mathbf{x}_i^l - \mathbf{x}_j^l \rVert^2,\,
a_{ij}
\right), \label{eq:egcl_edge} \\
\mathbf{x}_i^{l+1} &= \mathbf{x}_i^l
+ C \sum_{j \neq i} (\mathbf{x}_i^l - \mathbf{x}_j^l)\,
\phi_x(\mathbf{m}_{ij}), \label{eq:egcl_coord} \\
\mathbf{m}_i &= \sum_{j \neq i} \mathbf{m}_{ij}, \label{eq:egcl_agg} \\
\mathbf{h}_i^{l+1} &= \phi_h\!\left(
\mathbf{h}_i^l,\,
\mathbf{m}_i
\right). \label{eq:egcl_node}
\end{align}

\subsection{FLOPs Computation}
We largely follow (\cite{hoffmann2022trainingcomputeoptimallargelanguage}) with two modifications: (1) we replace the Mamba-2 block (\cite{gu2024mamba2}) with the Hydra quasiseparable mixer (\cite{hwang2024hydrabidirectionalstatespace}), and (2) we include the FLOPs introduced by the two SSD applications required for a quasiseparable (QS) mixer. A QS mixer decomposes into two SSDs plus a diagonal term:
\begin{equation}
\mathrm{QS}(X)=\operatorname{shift}(\mathrm{SS}(X))+\operatorname{flip}\!\left(\operatorname{shift}(\mathrm{SS}(\operatorname{flip}(X)))\right)+DX.
\end{equation}
Assuming the same $d_{\text{model}}$, $\text{expand}$, $d_{\text{state}}$, and $\text{num\_heads}$ as Mamba-2, the forward pass FLOPs for Hydra are:
\begin{itemize}
\item \textbf{XZ projections:} $2 \times \text{seq\_len} \times d_{\text{model}} \times (2 \times \text{expand} \times d_{\text{model}})$
\item \textbf{BC$\Delta t$ projections:} $2 \times \text{seq\_len} \times d_{\text{model}} \times (2 \times d_{\text{state}} + \text{num\_heads})$
\item \textbf{QS Mixer (two SSDs):}
\begin{itemize}
\item First SSD: $2 \times 3 \times \text{seq\_len} \times (\text{expand} \times d_{\text{model}}) \times d_{\text{state}}$
\item Second SSD: $2 \times 3 \times \text{seq\_len} \times (\text{expand} \times d_{\text{model}}) \times d_{\text{state}}$
\item Combined QS cost: $4 \times 3 \times \text{seq\_len} \times (\text{expand} \times d_{\text{model}}) \times d_{\text{state}}$
\end{itemize}
\item \textbf{Diagonal multiplication ($DX$):} $2 \times \text{seq\_len} \times d_{\text{model}}$
\item \textbf{Shift / Flip overhead:} $\approx 6 \times \text{seq\_len} \times d_{\text{model}}$
\item \textbf{Depthwise Convolution:} $2 \times \text{seq\_len} \times d_{\text{model}} \times \text{window\_size}$
\item \textbf{Gating:} $5 \times \text{seq\_len} \times d_{\text{model}}$
\item \textbf{Output projection:} $2 \times \text{seq\_len} \times d_{\text{model}}^{2}$
\end{itemize}
\subsection{Training \& Hyperparameters}
All models are trained using a batch size of 128 with a learning rate that peaks at $3 \times 10^{-4}$ and decays to a minimum of $3 \times 10^{-5}$. A warmup phase is applied, followed by a cosine learning rate schedule. Optimization is performed using Adam with a weight decay of 0.1 and momentum parameters $\beta_1 = 0.9$ and $\beta_2 = 0.95$. All experiments are conducted using the BF16 precision format.

For fairness, identical hyperparameter settings are used across all models without individual tuning, ensuring that any performance differences arise from architectural variations rather than optimization choices.

Training is performed using Distributed Data Parallel (DDP) with microbatch gradient accumulation. Each minibatch is divided into several microbatches, which are processed sequentially; gradients are accumulated locally using PyTorch's \texttt{no\_sync} mechanism and synchronized only on the final microbatch.

We employ a variance-preserving (VP) diffusion process with the standard cosine noise schedule proposed by \cite{nichol2021improveddenoisingdiffusionprobabilistic}. 
The cumulative product of signal coefficients is defined as
\begin{equation}
\bar{\alpha}_t = \frac{f(t)}{f(0)}, 
\qquad
f(t) = \cos^2\!\left( \frac{t/T + s}{1 + s} \cdot \frac{\pi}{2} \right),
\end{equation}
 where  \(s = 0.008\) is a small offset introduced for numerical stability.
The discrete noise variance at timestep \(t\) is then given by
\begin{equation}
\beta_t = 1 - \frac{\bar{\alpha}_t}{\bar{\alpha}_{t-1}},
\end{equation}
which ensures a smooth signal-to-noise ratio decay throughout the diffusion process.

\subsection{Distance Metrics}
\label{distance_metrics}
We adopt two distinct choices of distance metrics,outlined in Section \ref{inst}: the pseudo-Huber loss, which provides robustness to outliers during node-level denoising, and the graph-level mean squared error (MSE), which measures the overall reconstruction fidelity of the molecular graph structure.
The one-dimensional Pseudo-Huber loss is defined as
\begin{equation}
H_\delta(x) = \delta^2 \left( \sqrt{1 + \frac{x^2}{\delta^2}} - 1 \right).
\tag{12}
\end{equation}
In our implementation, we employ a normalized variant of the pseudo-Huber loss as the distance metric between predicted and target features. Specifically, for an input--target pair $(x, \hat{x})$, we define
\begin{equation}
\mathcal{L}_{\mathrm{PH}}(x, \hat{x}) = \sqrt{(x - \hat{x})^2 + \delta^2} - \delta,
\tag{10}
\end{equation}
where $\delta > 0$ controls the transition between the quadratic (L2-like) and linear (L1-like) regimes. This formulation is equivalent to the standard pseudo-Huber loss $H_\delta(x)$ in \cite{khrapov2024improvingdiffusionmodelssdatacorruption}.

\textbf{Graph-level MSE.} The graph-level mean squared error (MSE) quantifies the reconstruction quality of molecular graphs at the global (graph) level. Let $\mathcal{G} = \{G_1, \ldots, G_B\}$ denote a batch of $B$ molecular graphs, where each graph is represented as $G_g = (V_g, E_g)$ with node set $V_g = \{v_1, \ldots, v_{n_g}\}$ and edge set $E_g = \{e_{ij} \mid v_i, v_j \in V_g\}$. Across the batch, the total number of nodes is $N = \sum_{g=1}^{B} n_g$, with concatenated node features $X \in \mathbb{R}^{N \times d}$ and a batch assignment vector $b \in \{1, \ldots, B\}^N$, where $b_v = g$ indicates that node $v$ belongs to graph $G_g$. For each graph $G_g$, we compute its mean node feature vector as
\begin{equation}
\bar{x}_g = \frac{1}{n_g} \sum_{v \in V_g} x_v, \quad n_g = |V_g|.
\tag{10}
\end{equation}
The graph-level embeddings are aggregated into $\bar{X} = [\bar{x}_1; \ldots; \bar{x}_B] \in \mathbb{R}^{B \times d}$. Given the corresponding target graph-level representations $\hat{X} = [\hat{x}_1; \ldots; \hat{x}_B] \in \mathbb{R}^{B \times d}$, the graph-level MSE is defined as
\begin{equation}
\mathcal{L}_{\text{MSE-graph}} = \frac{1}{B} \sum_{g=1}^{B} \lVert \bar{x}_g - \hat{x}_g \rVert_2^2.
\tag{11}
\end{equation}
\subsection{Evaluation Metrics}
The core metrics used to evaluate generative models are validity, uniqueness, and diversity.

\textbf{Validity} quantifies the percentage of chemically sound structures within a generated set of molecules $G$. This is typically determined using molecular parsing tools like RDKit. Its calculation is defined as:
\begin{equation}
\text{Validity}(G) = \frac{|\{m \in G : \text{is\_valid}(m)\}|}{|G|} \times 100
\label{eq:validity}
\end{equation}
where $\text{is\_valid}(m)$ is a function that verifies the chemical validity of a molecule $m$.

\textbf{Uniqueness} measures the proportion of distinct, non-duplicate molecules in a generated set. It is computed as:
\begin{equation}
\text{Uniqueness}(G) = \frac{|\text{unique}(G)|}{|G|} \times 100
\label{eq:uniqueness}
\end{equation}
Here, $\text{unique}(G)$ returns the set of all unique molecules within $G$.

\textbf{Novelty}

Novelty measures the fraction of generated molecules that are not present in the training dataset $D_{\text{train}}$, indicating the model's ability to generate new, unseen molecules. For a generated set $G$, novelty is defined as:

\begin{equation}
\text{Novelty}(G) = \frac{|\{ m \in G : m \notin D_{\text{train}} \}|}{|G|} \times 100
\end{equation}

where $m$ represents a generated molecule and $D_{\text{train}}$ is the set of molecules seen during training. A higher novelty score indicates greater generation of previously unseen molecules.

\textbf{Diversity} assesses how well the generated set covers the chemical space. This is often quantified by the average pairwise Tanimoto distance between molecules' fingerprint representations. For a generated set $G$ and a fingerprint function $f(\cdot)$, diversity is given by:
\begin{equation}
\text{Diversity}(G) = \frac{2}{|G|(|G| - 1)} \sum_{i=1}^{|G|} \sum_{j=i+1}^{|G|} \text{Tanimoto}(f(m_i), f(m_j))
\label{eq:diversity}
\end{equation}
where $m_i, m_j \in G$ are molecules from the generated set, and $\text{Tanimoto}(\cdot, \cdot)$ calculates the Tanimoto similarity between their fingerprints.

\subsection{Quantitative Estimate of Druglikeness (QED)}
The Quantitative Estimate of Druglikeness (QED) is a widely adopted metric designed to quantify a molecule's druglikeness by combining eight key physicochemical properties into a single desirability score. 

The desirability function $d(p_i)$ for each individual property $p_i$ is empirically modeled using an asymmetric double sigmoidal function:
\begin{equation}
d(p_i) = \frac{1}{1 + \exp[-a_i(p_i - b_i)]} \times \frac{1}{1 + \exp[-c_i(p_i - d_i)]}
\label{eq:desirability}
\end{equation}
where $a_i, b_i, c_i, \text{ and } d_i$ are parameters fitted specifically for property $p_i$.

The overall QED value is then calculated by taking a weighted geometric mean of these individual desirability scores:
\begin{equation}
\text{QED} = \left( \prod_{i=1}^{n} d(p_i)^{w_i} \right)^{1 / \sum_{i=1}^{n} w_i}
\label{eq:qed}
\end{equation}
where $w_i$ represents the weight assigned to property $p_i$, reflecting its contribution to druglikeness.
QED scores range from 0 to 1, with higher values indicating greater druglikeness. This metric offers a continuous and nuanced assessment, proving more effective than binary, rule-based methods like Lipinski’s Rule of Five, particularly in identifying drug-like compounds that may not conform to traditional filters. It also facilitates the prioritization of promising drug candidates within specific chemical spaces.

\subsection{Atomic Stability}

Atomic stability measures whether individual atoms in a generated molecule satisfy fundamental chemical constraints, such as valency, charge balance, and bonding rules. For an atom $a \in m$, define:

\begin{equation}
\text{AtomicStability}(a) =
\begin{cases} 
1, & \text{if } \text{valence}(a) \le \text{max\_valence}(a) \text{ and charge constraints satisfied} \\
0, & \text{otherwise}
\end{cases}
\end{equation}

The overall atomic stability of a molecule $m$ is then:

\begin{equation}
\text{AtomicStability}(m) = \frac{\sum_{a \in m} \text{AtomicStability}(a)}{|m|}
\end{equation}

This yields a value in $[0,1]$, where 1 indicates that all atoms satisfy chemical stability rules.

\subsection{Molecular Stability}

Molecular stability evaluates the energetic and structural soundness of the entire molecule, considering criteria such as strain energy, bond length/angle deviations, or formal charge consistency. For a molecule $m$, a discrete definition is:

\begin{equation}
\text{MolecularStability}(m) =
\begin{cases} 
1, & \text{if all bonds and angles satisfy predefined chemical thresholds} \\
  & \text{and the molecule is free of high-energy conflicts} \\
0, & \text{otherwise}
\end{cases}
\end{equation}

Alternatively, a continuous molecular stability score can be computed using a normalized energy function $E(m)$:

\begin{equation}
\text{MolecularStability}(m) = \exp(-\alpha E(m))
\end{equation}

where $\alpha$ is a scaling factor. Values closer to 1 indicate higher molecular stability.

\subsection{3D Conformer Generation Metrics}

To evaluate the similarity between generated conformers and reference conformers, we employ two ensemble-based metrics: \emph{Average Minimum RMSD (AMR)} and \emph{Coverage}. Both metrics are reported in terms of \emph{Recall (R)} and \emph{Precision (P)}, capturing complementary aspects of conformer quality. Recall measures how well the generated ensemble spans the diversity of ground-truth conformers, while Precision quantifies how accurately the generated conformers match reference structures.

Let $\mathcal{G} = \{ g_1, \dots, g_{|\mathcal{G}|} \}$ denote the set of generated conformers, $\mathcal{R} = \{ r_1, \dots, r_{|\mathcal{R}|} \}$ denote the set of reference conformers, and $\mathrm{RMSD}(g, r)$ denote the root-mean-square deviation between conformers $g$ and $r$, computed after optimal alignment.

\paragraph{Average Minimum RMSD (AMR).}
The Average Minimum RMSD (AMR) measures the average distance between each conformer in one set and its closest match in the other set.

The \emph{Recall AMR} is defined as
\begin{equation}
\mathrm{AMR}_{\mathrm{R}}(\mathcal{G}, \mathcal{R}) =
\frac{1}{|\mathcal{R}|}
\sum_{r \in \mathcal{R}}
\min_{g \in \mathcal{G}}
\mathrm{RMSD}(r, g),
\end{equation}
which evaluates how well the generated conformers cover the diversity of reference conformers.

The \emph{Precision AMR} is defined as
\begin{equation}
\mathrm{AMR}_{\mathrm{P}}(\mathcal{G}, \mathcal{R}) =
\frac{1}{|\mathcal{G}|}
\sum_{g \in \mathcal{G}}
\min_{r \in \mathcal{R}}
\mathrm{RMSD}(g, r),
\end{equation}
which measures the geometric accuracy of the generated conformers. Lower AMR values indicate better performance.

\paragraph{Coverage.}
Coverage measures the fraction of conformers in one set that are matched within a predefined RMSD threshold $\delta$.

The \emph{Recall Coverage} is defined as
\begin{equation}
\mathrm{Coverage}_{\mathrm{R}}(\mathcal{G}, \mathcal{R}) =
\frac{1}{|\mathcal{R}|}
\sum_{r \in \mathcal{R}}
\mathbb{I}
\left[
\min_{g \in \mathcal{G}}
\mathrm{RMSD}(r, g) < \delta
\right],
\end{equation}
and the \emph{Precision Coverage} is defined as
\begin{equation}
\mathrm{Coverage}_{\mathrm{P}}(\mathcal{G}, \mathcal{R}) =
\frac{1}{|\mathcal{G}|}
\sum_{g \in \mathcal{G}}
\mathbb{I}
\left[
\min_{r \in \mathcal{R}}
\mathrm{RMSD}(g, r) < \delta
\right],
\end{equation}
where $\mathbb{I}[\cdot]$ denotes the indicator function. Coverage values range from $0$ to $1$, with higher values indicating better performance. In practice, $\delta$ is typically set to $0.5\,\text{\AA}$, $1.0\,\text{\AA}$, or $2.0\,\text{\AA}$, depending on the evaluation protocol.

\subsection{Limitations}
We design the choice of the depth and the scalable architecture for the SSMs or global attention module of our framework, following the theoretical and empirical results from Wang (2025) (\cite{wang2025understandingmitigatingbottlenecksstate}). Theorem 3.1 from Wang (2025) (\cite{wang2025understandingmitigatingbottlenecksstate}) states the dependencies among tokens in an SSM layer decay exponentially with distance, producing a localized representation effect. Due to the exponential decay of interactions between distant tokens, SSMs effectively behave as localized kernels, similar to the convolutional operators in CNNs or message-passing functions in GNNs. Goodfellow (\cite{goodfellow2016deep}) postulated that increasing the depth of a model is expected to expand its receptive field and improve its capacity to capture broader dependencies. We conduct our ablations by varying the number of layers and the model dimension.

\subsection{Detailed Model Specifications and Hyper-parameters.}
\textcolor{blue}{\textit{For fairness and reproducibility, we use an identical training setup across all models, keeping optimization, architectural, and diffusion hyperparameters unchanged to allow for a direct head-to-head comparison.}}
\begin{table}[htbp]
\vspace{-2pt}   
\centering
\small
\begin{tabular}{lcccc}
\toprule
\textbf{Attribute} & \textbf{Model 1} & \textbf{Model 2} & \textbf{Model 3} & \textbf{Model 4} \\
\midrule
Active Params & 63.0M & 63.4M & 71.7M & 55.5M \\
Total Params & 63.0M & 63.4M & 286.9M & 55.5M \\
\midrule
Layers & 8 & 8 & 8 & 8 \\
Channels & 512 & 512 & 512 & 512 \\
Depth & 6 & 6 & 6 & 6 \\
Head dim & 16 & 16 & 16 & 16 \\
Heads & 4 & 4 & 4 & 4 \\*
In-node-nf & 512 & 512 & 512 & 512 \\
Out-node-nf & 512 & 512 & 512 & 512 \\
In-edge-nf & 512 & 512 & 512 & 512 \\
Hidden-nf & 512 & 512 & 512 & 512 \\
Load Balancing Method & Loss-Free (\cite{wang2024auxiliarylossfreeloadbalancingstrategy}) & - & Loss-Free & - \\
\midrule
pe dim & 20 & 20 & 20 & 20 \\
context dim & 16 & 16 & 16 & 16 \\
time embed dim & 16 & 16 & 16 & 16 \\
Attention Variant & Transformer & Mamba /Mamba-2 & Jamba & Hydra \\
Position Embedding & RoPE (\cite{su2023roformerenhancedtransformerrotary}) & -- & RoPE & - \\
Layer Norm Type & RMSNorm (\cite{zhang2019rootmeansquarelayer}) & RMSNorm & RMSNorm & RMSNorm \\
\midrule
num tokens & 2000 & 2000 & 2000 & 2000 \\
Batch Size & 64 & 128 & 128 & 512 \\
Training Steps & 107500 & 107500 & 107500 & 107500 \\
Training graphs/molecules & 55k & 55k & 55k & 55k \\
Warmup Steps & 5000 & 5000 & 5000 & 5000 \\
\midrule
Optimizer & AdamW (\cite{loshchilov2019decoupledweightdecayregularization}) & AdamW & AdamW & AdamW \\
AdamW Betas & (0.9, 0.95) & (0.9, 0.95) & (0.9, 0.95) & (0.9, 0.95) \\
AdamW $\epsilon$ & $1\mathrm{e}{-20}$ & $1\mathrm{e}{-20}$ & $1\mathrm{e}{-20}$ & $1\mathrm{e}{-20}$ \\
Base Learning Rate & $3\mathrm{e}{-5}$ & $3\mathrm{e}{-5}$ & $3\mathrm{e}{-5}$ & $3\mathrm{e}{-5}$ \\
LR Scheduler & Cosine & Cosine & Cosine & Cosine \\
LR Decay Step Ratio & [0.8$\times$, 0.9$\times$] & [0.8$\times$, 0.9$\times$] & [0.8$\times$, 0.9$\times$] & [0.8$\times$, 0.9$\times$] \\
LR Decay Rate & [0.316, 0.1] & [0.316, 0.1] & [0.316, 0.1] & [0.316, 0.1] \\
Weight Decay & 0.1 & 0.1 & 0.1 & 0.1 \\
Loss-type & pseudo-huber & pseudo-huber & pseudo-huber & pseudo-huber \\
\bottomrule
\end{tabular}
\caption{Model architecture and detailed training hyperparameters}
\label{tab:mhc_hyperparams}
\end{table}

\end{document}